\documentclass[a4paper,fleqn]{cas-dc}
\usepackage{pifont}
\usepackage[numbers]{natbib}
\def\tsc#1{\csdef{#1}{\textsc{\lowercase{#1}}\xspace}}
\tsc{WGM}
\tsc{QE}
\tsc{EP}
\tsc{PMS}
\tsc{BEC}
\tsc{DE}

\begin{document}
\let\WriteBookmarks\relax
\def\floatpagepagefraction{1}
\def\textpagefraction{.001}

\shorttitle{Track-Guided Hierarchical Reinforcement Learning for Autonomous Vehicle Drifting with Minimum-Lap-Time Planning}

\shortauthors{Sheng Zhao et~al.}

\title [mode = title]{Track-Guided Hierarchical Reinforcement Learning for Autonomous Vehicle Drifting with Minimum-Lap-Time Planning}                      
\tnotemark[1]

\tnotetext[1]{This work was supported by the National Natural Science Foundation of China under Grant 52472413.}

\affiliation[1]{
    organization={School of Mechanical Engineering},
    organization={Shanghai Jiao Tong University},
    city={Shanghai},
    postcode={200240}, 
    country={China}}
\affiliation[2]{
    organization={School of Mechanical and Aerospace Engineering},
    organization={Nanyang Technological University},
    postcode={639798}, 
    country={Singapore}}

\author[1,2]{Sheng Zhao}\ead{zhaoshengkim@sjtu.edu.cn}
\credit{Writing - original draft preparation}

\author[2]{Bolin Zhao}\ead{bolin.zhao@ntu.edu.sg}
\credit{Writing - review \& editing}

\author[1]{Xiaodong Wu}\cormark[1]\ead{xiaodongwu@sjtu.edu.cn}
\credit{Writing - review \& editing, Conceptualization, Supervision}

\author[2]{Chen Lv}\cormark[1]\ead{lyuchen@ntu.edu.sg}
\credit{Conceptualization, Supervision}

\cortext[cor1]{Corresponding author}

\begin{abstract}
In Formula 1, drivers optimize racing lines within tire grip limits to minimize lap times; however, in rally racing, drivers intentionally break traction to drift on loose surfaces. This maneuver rapidly aligns the vehicle for corner exits, ultimately reducing lap time. Autonomously executing such maneuvers formulates a complex dual-objective control problem: stabilizing highly nonlinear drift dynamics while strictly minimizing lap time. Addressing this challenge motivates the development of advanced Minimum-Lap-Time (MLT) drift control architectures.
This paper proposes a planning-control framework specifically designed for MLT drifting scenario. First, we formulate an optimal control problem to generate a MLT drift planning trajectory, which is used as prior data to train a deep reinforcement learning drift controller. Given that drifting involves extremely large sideslip angles and is therefore challenging to learn directly, a Track-guided Reinforcement Learning (TgRL) drift control method is proposed to enable progressive training in a step-by-step manner—from drift control policy, to drift corner policy, and finally to a comprehensive drift race policy.
The reward function incorporates both an instant reward term and an end reward term derived from the Minimum-Lap-Time objective. Simulation results demonstrate that the proposed framework enables the agent to learn a drift racing policy that not only ensures vehicle motion control performance but also effectively reduces lap time.

\end{abstract}




\maketitle

\section{Introduction}
Drifting represents a boundary-condition maneuver in which the vehicle intentionally operates at or beyond the limits of its dynamic stability envelope \cite{Hindiyeh2014}. Conceptually, drifting can be regarded as a controllable loss of tire grip, where lateral forces are deliberately saturated to generate large sideslip angles. While such maneuvers enable novel possibilities in performance driving, they simultaneously impose substantial challenges on vehicle control due to strong nonlinearities in both tire and chassis dynamics \cite{Shan2024,Ni2017}. 

High-performance vehicle control on complex terrains, such as narrow race tracks with gravel, snow, or low-adhesion asphalt, remains highly challenging. Traditional cornering strategies often fail: aggressive maneuvers risk spin, while conservative ones waste time. Leveraging large sideslip angle (\(\beta\)) maneuvers, especially drifting, offers a promising alternative by expanding the vehicle’s dynamic envelope beyond the \(\pm 5^\circ\) range of steady-state cornering. This motivates a generalized notion of dynamic stability, where drifting is deliberately exploited. Accurate dynamic modeling under such limit-handling conditions must closely capture real-world system responses and provide a quantitative assessment of handling safety \cite{cheng2025}.

\subsection{Previous Work and Motivation}

Model-based drift control approaches typically rely on accurate system modeling and finely tuned control parameters tailored to specific vehicle configurations. However, the strong nonlinearities and parameter sensitivity inherent in vehicle–tire interactions often result in poor generalization across different platforms. To address these limitations, neural networks (NNs) have recently emerged as a data-driven paradigm capable of learning complex control strategies through trial-and-error interactions with the environment. For example, Spielberg \cite{Spielberg2019} employed NNs to approximate nonlinear vehicle dynamics for precise automated driving, while Broadbent \cite{Broadbent2024} modeled tire dynamics with NNs to achieve accurate automated drifting. Similarly, Weber \cite{Weber2023} proposed a modeling and control framework for dynamic drifting trajectories, and later extended this work to fit vehicle dynamics at large sideslip angles on low-friction roads, enabling NMPC-based trajectory generation \cite{Weber2024}. In parallel, reinforcement learning (RL) has demonstrated considerable potential in replicating expert-level driving behaviors—such as those of professional race car drivers—without requiring a complete analytical description of the underlying vehicle dynamics. \cite{ma2024} investigates autonomous vehicle performance under high-maneuvering conditions and large-curvature turns using a reinforcement learning method integrated with game theory. \cite{wu2023} explores eco-driving strategies for vehicles in traffic flow scenarios using deep reinforcement learning. \cite{Zhao_2025_drift} proposed a hybrid method that combines the advantages of RL and MPC to achieve sim-to-real drift experiments, demonstrating the significant potential of RL in the real world.

Notably, Sony's DRL-based agent \emph{Gran Turismo Sophy} surpassed professional drivers in competitive racing simulations, demonstrating advanced strategies such as slipstream overtaking and slingshot effects \cite{Wurman2021}. This landmark result, published in \textit{Nature}, confirmed the feasibility of DRL as a decision-making paradigm for high-performance driving. Besides this, Soft Actor–Critic (SAC) has played a pivotal role in autonomous control research, with demonstrated applications in trajectory following \cite{cai2020}, steady-state stabilization \cite{Jiang2021}, collision recovery \cite{Yin2020}, brake failure mitigation \cite{Zhao2024_Adaptive}, instability resolution via hybrid methods \cite{Toth2023}, and sim-to-real transfer \cite{Toth2024}. \cite{Wang2024} integrated a action mapping method within SAC frameworks and \cite{Hou2022} proposed an expert data-enhanced residual RL method. Proximal Policy Optimization (PPO) has shown strong performance in high-dimensional state spaces and vision-based drifting control \cite{Domberg2022}, modular vision-based driving \cite{Domberg2024}, and real-world precision control on a Toyota GR Supra \cite{Djeumou2025}. Twin Delayed Deep Deterministic Policy Gradient (TD3) has been employed to address precision requirements under coupled dynamics, for instance in tire grip loss scenarios \cite{Orgovan2021} and drift parking maneuvers \cite{Leng2023}. Meanwhile, Deep Q-Network (DQN) variants have been extended to incorporate feasibility in risk-sensitive tasks \cite{Bhattacharjee2018,Hoshino2024}. DRL has also emerged as a highly versatile approach for solving complex vehicle control problems, demonstrating remarkable success in applications ranging from energy-efficient eco-driving in uncertain urban traffic \cite{Li_2022_deep,Li_2023_deep} to the management of highly non-linear extreme vehicle dynamics. Collectively, these advancements highlight the adaptability of RL methods across diverse maneuvers, real-world deployment contexts, and extreme scenarios, thereby establishing a strong foundation for limit-handling autonomous systems.

\begin{figure}
    \centering
    \includegraphics[width=1\columnwidth]{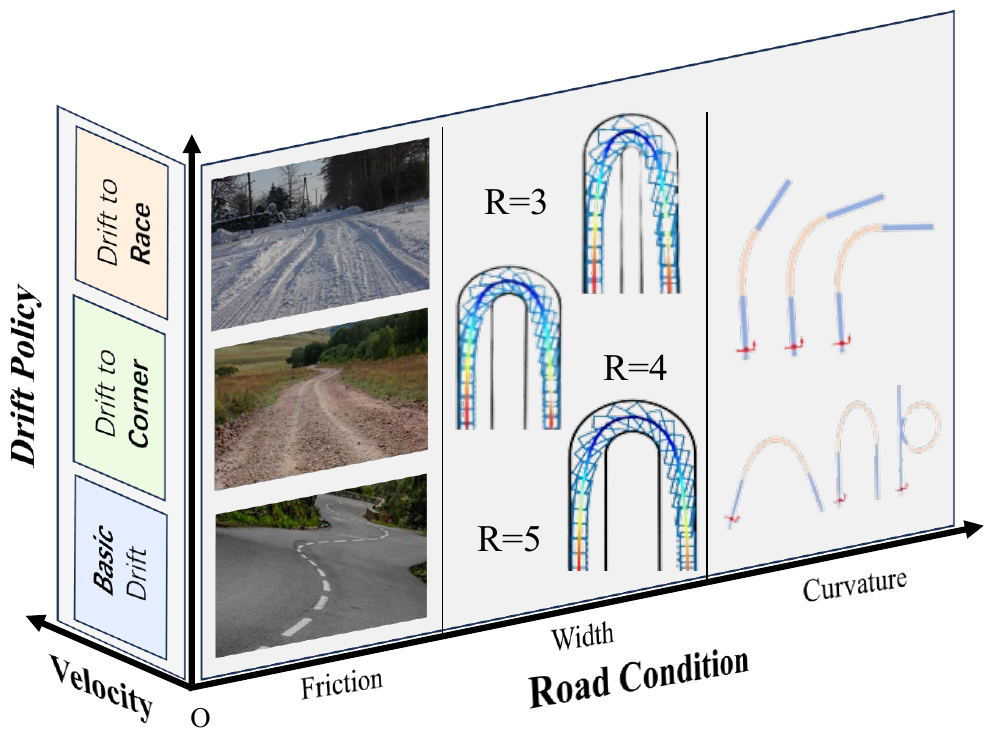}
    \caption{Essence of drift strategy: under varying road conditions (corner curvature, surface friction, lane width) and different vehicle speeds, the agent leverages the RL policy network to infer the corresponding optimal drift posture and control method.}
    \label{intro}
\end{figure}

As mentioned, Cai's RL method \cite{cai2020} closely resembles imitation learning, as it relies heavily on shaping reward function to imitate expert-level control actions. 
This reveals two central challenges faced by existing RL-based drift controllers. First, they typically rely on large-scale expert demonstration datasets that provide action-level supervision (steering and throttle inputs). Such datasets are costly, time-consuming, and highly vehicle-specific, thereby hindering scalability. Second, most approaches lack structured mechanisms to guide agents toward physically meaningful drift strategies during early training. Without informative priors, RL agents require extensive exploration, often converging to suboptimal or unsafe policies. To address these issues, we propose a novel training paradigm grounded in the intrinsic nature of drifting, where the agent’s drift policy is progressively guided according to road conditions. This paradigm encourages the agent to acquire the fundamental essence of drift strategies: namely, under varying conditions (surface friction, lane width, corner curvature), the agent should be able to infer the optimal drift posture and control strategy directly through the policy network, as illustrated in Fig.~\ref{intro}.

Drifting maneuvers commonly arise in two categories of scenarios: (i) extreme obstacle-avoidance situations \cite{LI2023,Bertipaglia_2024_model,zhao2024_Autonomous}, and (ii) racing scenarios such as the World Rally Championship (WRC) \cite{Hou2022,Kabzan2019,Djeumou2024}. In extreme avoidance cases, drivers typically prioritize braking to prevent rear-end collisions (e.g., at red-light intersections). Only when the vehicle is traveling at high speeds do drivers resort to rapid steering corrections to avoid collisions. According to the phase-plane analysis of vehicle dynamics \cite{huang2020,Lu2025}, large and abrupt steering inputs at high speeds often induce significant instability. In this context, drift control provides a means to confine the vehicle’s instability within a controllable, near-critical stability region, thereby enhancing overall safety \cite{LI2023}.

In racing scenarios, drifting is primarily employed to improve vehicle maneuverability during cornering. Professional race drivers often coordinate the use of the handbrake, throttle, brake pedal, and steering wheel to initiate and sustain drift maneuvers. By adjusting the vehicle’s heading angle without reducing speed, they maintain higher cornering velocities and thus minimize lap time. Motivated by this, several researchers have investigated drift-based racing strategies. For example, \cite{cai2020} developed a reinforcement-learning-based training approach across six different tracks, emphasizing the dynamic characteristics of high-speed drifting. Likewise, \cite{AJANOVIC2023} proposed a search-based drift planning method that requires precise road curvature information, with the objective of achieving the shortest possible lap time. The minimum-lap-time racing problem has also been extensively studied, with \cite{Hou2022} addressing it under tire friction limits.

However, in WRC or other uneven, low-adhesion asphalt conditions, tires are often unable to achieve their maximum grip potential, limiting the effectiveness of conventional racing lines. To address this, we consider designing a drift-based control strategy that minimizes lap time under low-friction track conditions. Our objective is explicit: to achieve the shortest lap time. The proposed Minimum-Lap-Time (MLT) drift planning method contributes in two key aspects: (1) it generates a drift trajectory that minimizes lap time by leveraging a high-fidelity vehicle model to compute dynamically feasible drift trajectories, including vehicle position and attitude states; and (2) it employs these trajectories as reference data within an RL controller, thereby eliminating the need for costly expert demonstration data from professional drivers.

\subsection{Contributions of This Work}
Building upon the above motivation, this paper makes the following contributions:

(1) This paper proposes a novel Track-guided Reinforcement Learning (TgRL) planning-control hierarchical framework tailored for Minimum-Lap-Time (MLT) drifting. To address the complex dual-objective of stabilizing highly nonlinear drift dynamics while strictly minimizing lap times, the framework seamlessly integrates an optimal-control-based trajectory planner with a deep reinforcement learning controller, enabling the autonomous agent to effectively execute and exploit extreme drift maneuvers.

(2) We formulate a Minimum-Lap-Time (MLT) drift planning method based on optimal control. This module generates dynamically feasible, aggressive reference trajectories that serve as high-quality prior data, effectively encapsulating the MLT objective and replacing the need for high-cost expert demonstrations to guide the RL controller.

(3) We introduce an innovative training methodology that integrates curriculum learning principles into the TgRL framework. To overcome the inherent challenges of directly learning extreme dynamics with exceptionally large sideslip angles, the vehicle agent is trained in a progressive, staged manner—advancing systematically from a fundamental drift control policy, to a drift cornering policy, and ultimately to a comprehensive drift race policy. This step-by-step approach is reinforced by a specialized reward function combining instant tracking terms and MLT-derived end rewards.

The remainder of this paper is organized as follows. Section~\ref{sec:modelling} introduces the required models. Section~\ref{sec:planning} presents the proposed MLT-based drift planning method. Section~\ref{sec:control} describes the control framework and the Track-guided RL training method and definition. Section~\ref{sec:validation} reports all setup and evaluations. Finally, Section~\ref{sec:conclusion} summarizes the findings and discusses future research directions.

\section{Modelling}\label{sec:modelling}
\subsection{Kinematics and Dynamics Vehicle Model}

\begin{figure}
    \centering
    \includegraphics[width=1\columnwidth]{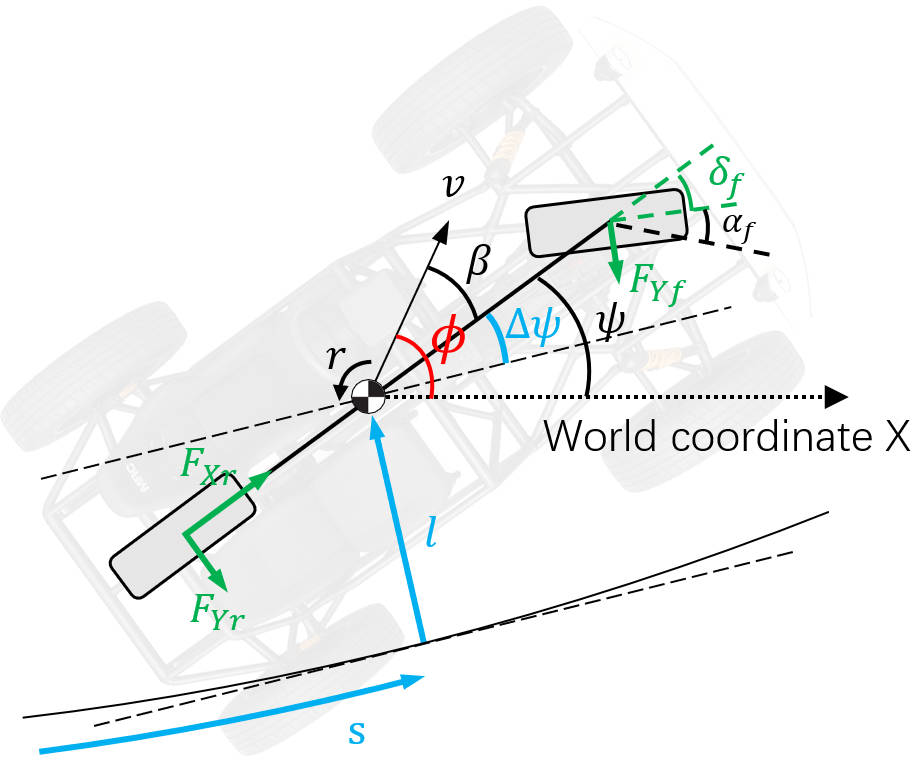}
    \caption{3-DoF vehicle kinematics and dynamics model for drift planning.}
    \label{model}
\end{figure}

A nonlinear 3-DoF dynamic model is derived based on the single-track "bicycle" configuration, as illustrated in Fig~\ref{model}. The equations of motion are then
\begin{equation}
\left\{
\begin{aligned}
\dot{v} &= \dfrac{ - F_{Y_f} \sin(\delta - \beta) + F_{X_r} \cos(\beta) + F_{Y_r} \sin(\beta)}{m} \\
\dot{\beta} &= \dfrac{ F_{Y_f} \cos(\delta - \beta) - F_{X_r} \sin(\beta) + F_{Y_r} \cos(\beta)}{mv} - r \\
\dot{r} &= \dfrac{a F_{Y_f} \cos(\delta)  - b F_{Y_r}}{I_z}
\end{aligned}
\right.
\end{equation}

To enable reference-path-relative control and trajectory planning, the vehicle dynamics are further expressed in the Frenet coordinate system, where the states are decomposed into longitudinal and lateral deviations from the reference path.
The Frenet coordinate system provides a more precise representation of the vehicle's motion relative to the reference path and facilitates the simplification of path tracking controller design. The corresponding kinematic model can be expressed as:
\begin{equation}
\left\{
\begin{aligned}
\dot{s} &= \dfrac{V \cos(\Delta\psi + \beta)}{1 - l \cdot \kappa} \\
\dot{l} &= V \sin(\Delta\psi + \beta) \\
\Delta\dot{\psi} &= r - \kappa \dot{s}
\end{aligned}
\right.
\label{kinematic modelling}
\end{equation}

\subsection{Tire Model}

During extreme drifting maneuvers, tires often operate beyond the linear region and enter saturation, resulting in complex conditions characterized by coupled longitudinal slip and lateral sideslip. Accurately modeling the tire dynamics under such conditions is critical. The Magic Formula tire model, which is derived from extensive experimental data fitting, is widely adopted in the vehicle simulation industry. Its primary advantages lie in its high fitting accuracy, strong parameter adaptability, and suitability for real-time applications. Therefore, this study employs the Magic Formula under combined slip conditions to model tire behavior.
The tire slip angle $\alpha_f$ and slip ratio $\lambda_r$ are represented as Eq.~\eqref{alpha_f} and \eqref{lambda}.

\begin{equation}
a_f = \arctan\left(\frac{v \sin \beta + a r}{v \cos \beta}\right) - \delta_f
\label{alpha_f}
\end{equation}
\begin{equation}
\lambda_r = \dfrac{v_x - \omega \cdot r_e}{v_x}
\label{lambda}
\end{equation}

The kinematic and dynamic analysis of the vehicle is conducted using a simplified model, with the equations derived based on the combined longitudinal and lateral forces acting on the vehicle. The longitudinal, lateral tire slip \( s_x,s_y \) and combined slip \( s \) are expressed as follows:

\begin{equation}
\left\{
\begin{aligned}
s_x &= \frac{\lambda_r}{1 + \lambda_r} \\
s_y &= \frac{\tan(\alpha_f)}{1 + \lambda_r}\\
s &= \sqrt{s_x^2 + s_y^2}
\end{aligned}
\right.
\label{eq:kinematics}
\end{equation}

The forces acting on the vehicle are defined in the following system:
\begin{equation}
\left\{
\begin{aligned}
\mu &= D \sin(C \arctan((1 - E) B s_x + E \arctan(B s_x))) \\
F_X &= \frac{s_x}{s} \mu F_z \\
F_Y &= \frac{s_y}{s} \mu F_z
\end{aligned}
\right.
\label{eq:forces}
\end{equation}

Here, \( \mu \) represents the friction coefficient, \( F_X \) and \( F_Y \) denote the longitudinal and lateral forces, respectively, and \( F_z \) is the vertical load. Parameters \( B \), \( C \), \( D \), and \( E \) are empirically determined coefficients.
A low road friction coefficient is used to simulate slippery surfaces where the tire forces cannot be fully utilized.

The vehicle and tire models developed in this section are applied in the drift planning described in Section~\ref{sec:planning}. 
All parameters and variables used in the model are summarized in Table~\ref{canshu1}.

\begin{table}[t!]
\caption{\textbf{Parameter explanation.}}
\label{canshu1}
\centering
\resizebox{\columnwidth}{!}{%
\begin{tabular}{@{} lcc @{}} 
\toprule
\textbf{Symbol}&\textbf{Unit}&\textbf{Value} \\
\midrule
Vehicle mass $m$ &Kg& 1300 \\
CG to front axle $a$ &m&1.4 \\
CG to rear axle $b$ &m&1.4 \\
Inertia moment $I_z$ &rad&1750 \\
Tire radius $r_e$ &m&0.32 \\
Road friction coefficient $\mu$ & - &0.6\\
course angle $\phi$ &rad&- \\
heading/yaw angle $\psi$ &rad&- \\
heading/yaw angle error to reference $\Delta \psi$ &rad&- \\
Longitudinal velocity $v_x$ &m/s&- \\
Lateral velocity $v_y$ &m/s&- \\
Combined velocity $v$ &m/s&- \\
Yaw rate  $r$ &rad/s&- \\
Sideslip angle $\beta$ & rad &-\\
Trajectory offset $e_y$ & m &-\\
heading/yaw angle error $e_{\psi}$ & rad &-\\
Sideslip angle error $e_{\beta}$ & rad &-\\
Vehicle location $(x,y)$ & - &-\\
Reference road curvature $\kappa_{ref}$ & 1/m &-\\
Front wheel steering angle $\delta_{f}$ & rad &-\\
Hand wheel steering angle $\delta$ & rad &-\\
Lateral tire force $F_Y$ & N &-\\
Longitudinal tire force $F_X$ & N &-\\
Lateral displacement in Frenet coordinate $l$ & m &-\\
Longitudinal displacement in Frenet coordinate $s$ & m &-\\
Rear tire wheel rotation speed $\omega$ & rpm &-\\
Front tire slip angle $\alpha_f$ & rad &-\\
\bottomrule
\end{tabular}%
}
\end{table}

\section{Drift Planning for Minimum lap time }\label{sec:planning}
Traditional trajectory planning algorithms focused on vehicle steady-state behavior typically aim to improve ride comfort, energy efficiency, and traffic efficiency \cite{Li2025}. However, under extreme driving conditions such as drifting through sharp turns, conventional methods often fail to generate safe and feasible trajectories that meet the demands of emergency scenarios.

In contrast, drift trajectory planning on racetracks considers vehicle dynamics, tire characteristics, and road constraints to generate trajectories that can be followed using controlled drifting maneuvers. Unlike traditional steady-state trajectory planning, which emphasizes long-term stability, drift trajectory planning prioritizes the vehicle's instantaneous maneuverability. This section presents a planning method for racetrack drifting considering the lap time.
\subsection{Minimum-Lap-Time Drifting Planning based on OCP}

We build the drift planning as a optimal control problem. The general form of the optimal control problem is given by:

\begin{equation}
\min J = \Phi(x(t_0), t_0, x(t_f), t_f) + \int_{t_0}^{t_f} L(x(t), u(t), t) \, dt
\end{equation}
\begin{equation}
\text{s.t.} \quad \frac{dx}{dt} = f(x(t), u(t), t)
\label{state variables}
\end{equation}
\begin{equation}
g(x(t), u(t), t) \leq 0
\label{budengshi}
\end{equation}
\begin{equation}
h(x(t), u(t), t) = 0
\label{dengshi}
\end{equation}

In the above:
\begin{itemize}
    \item \( J \) is the performance index.
    \item \( \Phi(x(t_0), t_0, x(t_f), t_f) \) is the terminal cost, representing the cost associated with the initial and final states.
    \item \( L(x(t), u(t), t) \) is the running cost, representing the continuous performance index over time.
    \item \( x(t) \) denotes the state variables of the system.
    \item \( u(t) \) denotes the control inputs.
    \item \( t_0 \) and \( t_f \) are the initial and final times, respectively.
    \item \eqref{state variables} describes the system dynamics in the form of ordinary differential equations.
    \item \eqref{budengshi} and \eqref{dengshi} define the inequality and equality constraints, respectively.
\end{itemize}

These constraints are used to impose limits throughout the entire trajectory in the optimal control problem (OCP), such as road boundaries, state constraints, and control input limitations.
We define the state variable and control input of the system model are as follows.
\[
\mathbf{x} = \begin{bmatrix} v \\ \beta \\ r \\ l \\ \Delta \psi \end{bmatrix}, \quad
\mathbf{u} = \begin{bmatrix} \delta_f \\ \lambda_r  \end{bmatrix}.
\]

After coordinate transformation, the forces acting on the vehicle's CG in the body coordinate system are analyzed, excluding external forces. Based on the simplified vehicle model, the final system dynamics equation is obtained as below according to the s,
\begin{equation}
\label{state_equ}
    \begin{bmatrix}
{v}' \\
\beta' \\
{r}' \\
{l}'\\
{\Delta \psi}'
\end{bmatrix}
= \frac{1}{\dot{s}} \begin{bmatrix}
- F_{Y_f} \sin(\delta_f - \beta) + F_{X_r} \cos(\beta) + F_{Y_r} \sin(\beta) \\
 F_{Y_f} \cos(\delta_f - \beta) - F_{X_r} \sin(\beta) + F_{Y_r} \cos(\beta) \\
a F_{Y_f} \cos(\delta_f) - b F_{Y_r} \\
v \sin(\Delta \psi + \beta)\\
r-\kappa\dot{s}
\end{bmatrix} .
\end{equation}

\begin{figure}
    \centering
    \includegraphics[width=1\columnwidth]{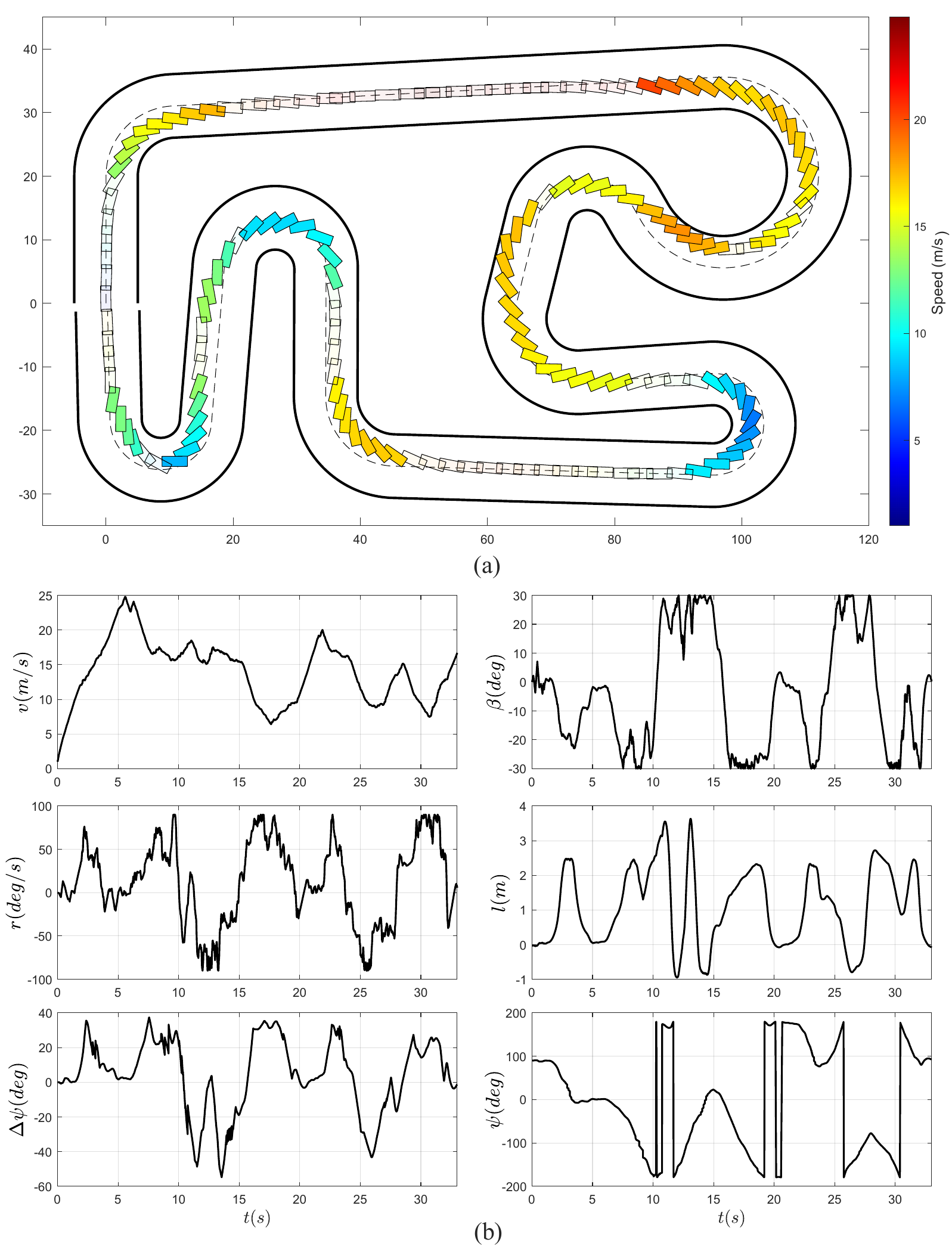}
    \caption{MLTDP planning result example on Lkarting track: (a) MLT trajectory. In the diagram, the vehicle box is colored based on the vehicle speed. When the sideslip angle is less than 15°, the vehicle box is given transparency to better highlight when the vehicle is in a drifting state; (b) MLT state variables by planning method.}
    \label{planning}
\end{figure}

Through Minimum-Lap-Time Drifting Planning (MLTDP), we can obtain a race trajectory enriched with state information, including trajectory coordinates and their corresponding vehicle speed, sideslip angle, yaw rate, and course angle. An example of the optimized planning result is shown in Fig~\ref{planning}. These data serve as crucial prior information for the drift tracking control discussed in Section~\ref{sec:control}.

\subsection{Objective function}

The problem of planning the MLT trajectory through a track for a vehicle can be formulated as an OCP. In this formulation, the performance index of the optimal control problem is the minimum time to traverse the curve:

\begin{equation}
\min J = \int_{t_0}^{t_f} \, dt
\end{equation}
where, \( t_0 \) is known, \( t_f \) is unknown, so it belongs to the class of optimization problems with a fixed initial point and a free terminal point, which is not easy to solve directly. Therefore, it is advantageous to transform to the Frenet coordinate system. As shown in Eq~\eqref{frenettrans}, the time integral is converted into an arc-length integral. In this transformation, the optimization objective becomes to minimize the time to pass through the arc-length segment from \( s_0 \) to \( s_f \). As a result, fixing \( s_0 \) and \( s_f \) converts it into a fixed-terminal optimal control problem:

\begin{equation}
\min J = \int_{t_0}^{t_f}  \, dt = \int_{s_0}^{s_f} \frac{dt}{ds} \, ds = \int_{s_0}^{s_f} \frac{1}{\dot{s}} \, ds.
\label{frenettrans}
\end{equation}

From kinematic modelling \eqref{kinematic modelling}, we substitute

\begin{equation}
\dot{s} = \frac{ds}{dt} = \frac{v\cdot \cos(\Delta \psi + \beta)}{1 - l \cdot \kappa_{ref}}
\end{equation}
into \eqref{frenettrans}, so we can obtain the final performance index:
\begin{equation}
\min J = \int_{s_0}^{s_f} \frac{1 - l \cdot\kappa_{ref}}{v\cdot \cos(\Delta \psi + \beta)} \, ds.
\end{equation}

\subsection{Multiple constraints}

Besides the state-space constraints as Eq~\eqref{state variables} and \eqref{state_equ}, the system must also satisfy the following additional conditions.

\subsubsection{State variable constraints}

The State variable constraint is according to the dynamics vehicle system, which is given by the following system of inequalities.
\begin{equation}
\begin{cases}
0 \leq v \leq 30 \\
-{\pi}/{6} \leq \beta \leq {\pi}/{6} \\
-\frac{\pi}{2} \leq r \leq \frac{\pi}{2} \\
-\frac{\pi}{2} \leq \Delta \phi \leq \frac{\pi}{2} \\
-l_0 \leq l \leq l_0
\end{cases}
\end{equation}
\subsubsection{Control variable constraints}
Eq~\eqref{Control variables} defines the bounds on the control variables. Based on engineering experience, the front wheel steering angle \( \delta_f \) is typically limited to within \( \pm 30^\circ \). The slip ratio of the driving wheels is required to be greater than \(-1\).

\begin{equation}
\begin{cases}
-{\pi}/{6} \leq \delta \leq {\pi}/{6} \\
-1 \leq \lambda_r
\end{cases}
\label{Control variables}
\end{equation}

\subsubsection{Drivable boundary constraints}

\begin{figure}
    \centering
    \includegraphics[width=0.6\columnwidth]{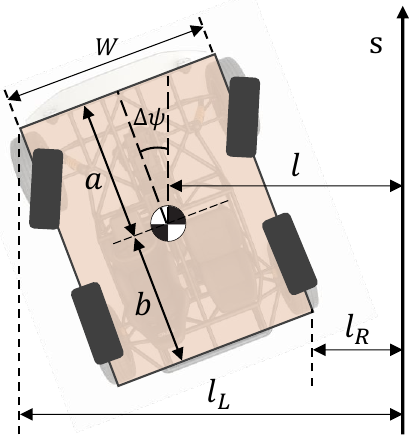}
    \caption{Drivable boundary constraints definition.}
    \label{drivable boundary}
\end{figure}

Fig~\ref{drivable boundary} and Eq~\eqref{Drivable roads} represent the road boundary constraints.
\begin{equation}
\begin{cases}
-l_0 \leq l_L \leq l_0 \\
-l_0 \leq l_R \leq l_0
\end{cases}
\label{Drivable roads}
\end{equation}
where $l_L$ represents the lateral position $l$ value at the starting point of the vehicle turn, and $l_R$ represents the lateral position $l$ value at the endpoint of the vehicle. From the frenet modelling, it can be seen that the two lateral displacement can be calculated considered to course angle deviation. 
\begin{equation}
\begin{cases}
l_L = l + \sqrt{a^2 + \left(\dfrac{W}{2}\right)^2}  \sin\left(\Delta \psi + \arctan \dfrac{W}{2a}\right)\\
l_R = l - \sqrt{b^2 + \left(\dfrac{W}{2}\right)^2}  \sin\left(\Delta \psi + \arctan \dfrac{W}{2b}\right)
\end{cases}
\end{equation}
Where: $l$ and $s$ is the frenet coordinate value at the CG, $a$ is the distance from CG to the front axle (assume that the vehicle length is equal to wheelbase), $b$ is the distance from CG to the rear axle, $W$ is the vehicle width, and $\Delta \psi$ is the heading angle deviation.

\subsubsection{Endpoint constraints}

The initial and final value constraints for the entire optimization problem are shown in Table \ref{endpoint_const}. The positions of the vehicle and the corner are not set with initial and final values, meaning that this optimization problem can solve for the optimal driving trajectory and initial position under the current path.

\begin{table}[H]
\caption{Endpoint constraints for fastest cornering planning}
\centering
\begin{tabular}{ccc}
\toprule
 \textbf{Variable} & \textbf{Initial Value} & \textbf{Final Value} \\
\midrule
 $v$ & Free & Free \\
$\beta$ & $\beta_0 = 0$ & $\beta_f = 0$ \\
$r$ & $r_0 = 0$ & $r_f = 0$ \\
$l$ & Free & Free \\
 $\Delta \psi$ & $\Delta \psi_0 = 0$ & $\Delta \psi_f = 0$ \\
 $s$ & $s_0 = 0$ & $s_f = \text{const}$ \\
\bottomrule
\end{tabular} \label{endpoint_const}
\end{table}

\section{Racing Drift Control with track-guided \\Deep reinforcement learning}\label{sec:control}
\subsection{Framework}
\begin{figure*}
    \centering
    \includegraphics[width=1\textwidth]{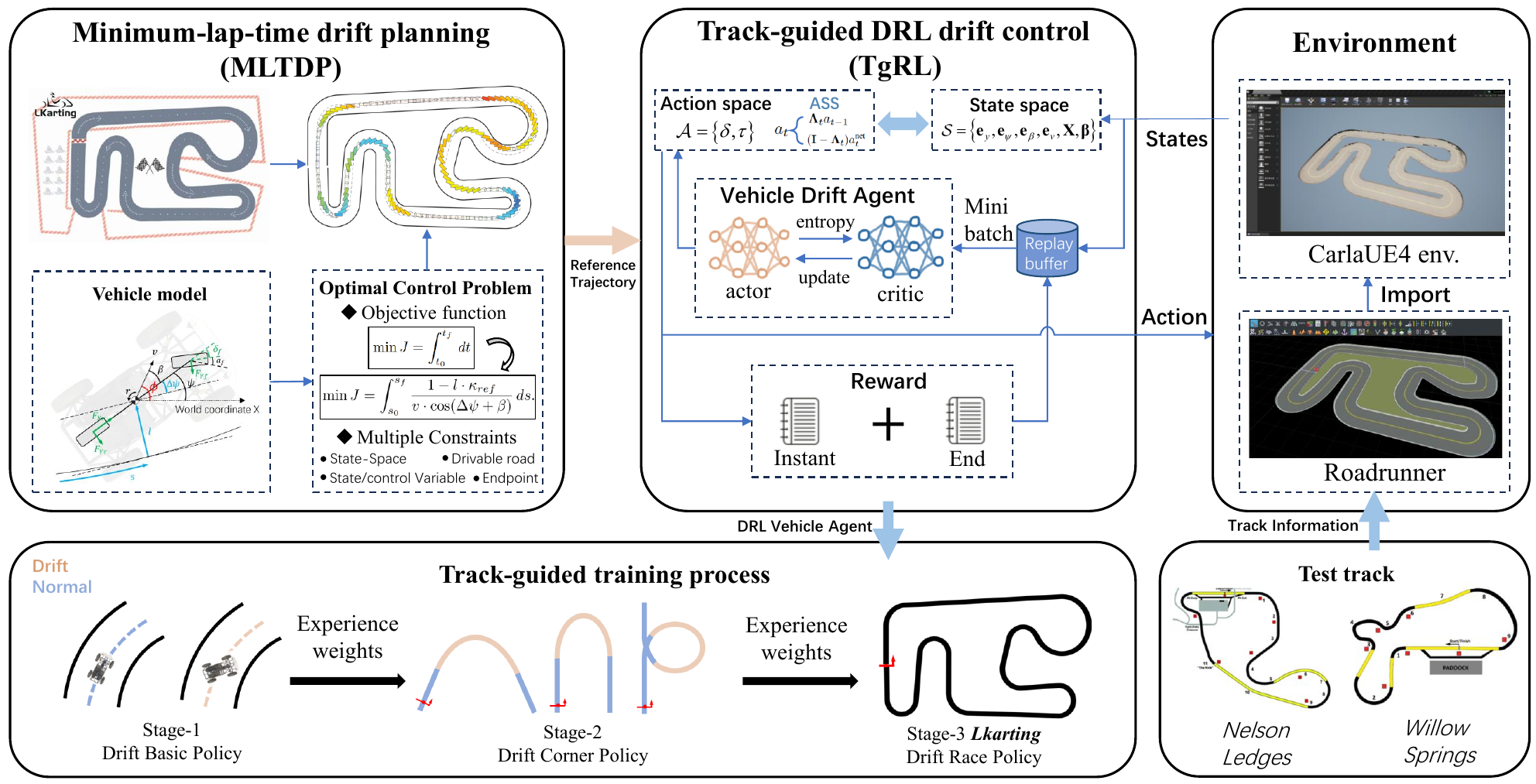}
    \caption{The proposed MLTDP-TgRL vehicle autonomous drifting framework.}
    \label{framework}
\end{figure*}
We propose a DRL training framework for autonomous drifting based on optimal racing trajectories, shown in Fig~\ref{framework}. The framework consists of two key components: Minimum-Lap-Time Drift Planning (MLTDP) and Track-guided Reinforcement Learning (TgRL) Drift Control. 

The MLTDP module formulates an optimization problem grounded in a high-fidelity vehicle dynamics model. Road boundaries and dynamic constraints are incorporated as constraints in the optimization, with the objective being to minimize lap time. Although this approach can yield high-performance drift trajectories, it heavily relies on accurate dynamic models, which are often difficult and expensive to obtain in real-world applications, whether for planning or control. To address this challenge, we leverage DRL for drift control, utilizing neural networks to capture the nonlinearities inherent in vehicle dynamics. This allows us to design a controller that does not require an explicit dynamic model.

The TgRL via DRL module is trained on a variety of track types and evaluated on a real-world race park. TgRL exploits the prior drifting data generated by MLTDP to guide the vehicle agent in learning optimal cornering trajectories and control policies. A step-by-step training strategy guided by track information is employed in TgRL, which significantly facilitates more effective learning of the complex drifting maneuver, also improves training efficiency and convergence speed. Details of the track-guided training method are presented in Section~\ref{Track_guided}.

\subsection{State Space}
Based on the state variables introduced above, the state space \( s \in \mathcal{S} \) is defined as Eq~\eqref{statespace},
\begin{equation}
    \mathcal{S} = \left\{  \mathbf{E}_y, \mathbf{E}_\psi, \mathbf{E_\beta}, \mathbf{{E}}_v,\boldsymbol{\beta} ^N,\mathbf{X}^N\right\},  \label{statespace}
\end{equation}

where \( \mathbf{X}^N \) and \(\boldsymbol{\beta}^N\) contain N location points \((x, y)\) and N sideslip angles \(\beta\) in the reference trajectory ahead.  \( e_\beta \) is the slip angle difference between the vehicle and the reference trajectory. \( e_{v} \) is the error of the velocity. To simplify notation, we define the following vectors:
\[
\mathbf{E}_y = \begin{bmatrix} e_y \\ \dot{e}_y \end{bmatrix}, \quad
\mathbf{E}_\psi = \begin{bmatrix} e_\psi \\ \dot{e}_\psi \end{bmatrix}, \quad
\mathbf{E}_\beta = \begin{bmatrix} e_\beta \\ \dot{e}_\beta \end{bmatrix}, \quad
\mathbf{E}_v = \begin{bmatrix} e_v \\ \dot{e}_v \end{bmatrix}.
\]

Therefore, the dimension of the set \( \mathcal{S} \) is determined by the parameter \( N \),  
which typically takes the value \( N = 10 \).

To accommodate termination conditions during training, a terminal state indicator, \textit{EndFlag}, is introduced. A training episode is terminated once the vehicle experiences a collision, deviates by 180° from the reference heading (i.e., performs a U-turn), or reaches the predefined goal.

\subsection{Action Space}
The continuous action space \( a \in \mathcal{A} \) is defined as Eq~\eqref{actionspace},
\begin{equation}
\mathcal{A} = \left\{ \delta, \tau \right\} \label{actionspace}
\end{equation}
where \( \delta \) is the steering angle and \( \tau \) is the throttle input. These control variables are normalized to \( \delta \in [-1, 1] \) and \( \tau \in [0, 1] \), respectively. Because we focus more on lap time and there will be no competing vehicles in the testing scenario, the braking control variable will be omitted and the throttle input is constrained to \( \tau \in [0.6, 1] \), thereby avoiding undesirably slow driving and enhancing training efficiency. Furthermore, to mitigate the risk of rollover caused by aggressive steering at high speeds \cite{cai2020}, the steering angle is limited to a narrower range of \( \delta \in [-0.8, 0.8] \).

To further improve control stability under varying vehicle speeds, we propose an Adaptive Action Smoothing (AAS) strategy. The smoothed action at time step \( t \) is computed as Eq~\eqref{action} to handle the multi-dimensional nature of the control action \( a_t = [\delta_t,\ \tau_t]^\top \).

\begin{equation}
    a_t = \boldsymbol{\Lambda}_t a_{t-1} + (\mathbf{I} - \boldsymbol{\Lambda}_t) a_t^{\text{net}},\label{action}
\end{equation}
where \( \boldsymbol{\Lambda}_t = \mathrm{diag}(\lambda_{\delta,t}, \lambda_{\tau,t}) \in \mathbb{R}^{2 \times 2} \) determines the individual smoothing weights for steering and throttle. \( a_t^{\text{net}} \) is the action predicted by the policy network at time \( t \), and \( a_{t-1} \) is the previously deployed action.

The steering smoothing factor \( \lambda_{\delta,t} \) is adapted based on the current speed \( v_t \) as:
\begin{equation}
    \lambda_{\delta,t} = \lambda_{\delta,\min} + (\lambda_{\delta,\max} - \lambda_{\delta,\min}) \cdot \dfrac{v_t - v_{\min}}{v_{\max} - v_{\min}},
\end{equation}
while \( \lambda_{\tau,t} \) is set to a small constant (e.g., \( 0.2 \)) to retain responsiveness in throttle control during drifting. This design ensures that steering commands become smoother at higher speeds to reduce rollover risk, while throttle remains responsive for effective acceleration.

The drift control agent trained with AAS is capable of exploring aggressive drifting maneuvers while maintaining vehicle stability and preventing rollover. This leads to improved tracking performance of the overall drift control system.

\subsection{Reward Function}
Multi-objective reward functions provide a robust framework for optimization in reinforcement learning \cite{Li_2023_overcoming,Li_2024_multiobjective}. By carefully shaping the reward, we can accelerate convergence and ensure the RL agent learns behaviors consistent with human design. We propose a combined reward strategy utilizing both instantaneous and terminal rewards. This approach systematically evaluates the vehicle's drifting process through kinematic and dynamic metrics, supplemented by explicit success and failure criteria.

\subsubsection{Instant Rewards} 
Instant rewards refer to the reward signals assigned to the agent at every time step within an episode. To guide the agent toward learning drift behavior, we define a reward mechanism that positively reinforces high-speed cornering with controlled large sideslip angles—a characteristic of effective drifting. When the vehicle enters a high sideslip state, it is considered to be drifting, and a corresponding positive reward is granted to encourage this behavior.

A well-defined reward function is critical for evaluating the controller’s performance in the context of high-speed drifting. The main control objectives are to minimize the lateral deviation error \( e_y \), heading angle error \( e_\psi \), and sideslip angle error \( e_\beta \) \cite{cai2020}. Besides these, we add the velocity error term \( e_v \) to formulate instant reward terms \( r_{e_y}, r_{e_\psi}, r_{e_\beta},r_{e_v} \), which quantify the agent’s performance with respect to each respective metric. $r_{e_y}$ has slowed down the rate of reward decline. For the reward function design of $r_{e_\psi}$ and $r_{e _\beta}$, we adopted a Gaussian distribution function. This reward function configuration can slow down the decline rate of rewards in intervals with small errors and quickly penalize in intervals with large errors. This combined reward function allows the vehicle agent to deviate from the planned trajectory for exploration, but it also needs to follow the course angle and sideslip angle.  These reward components are defined as Eq~\eqref{eq:reward}.

\begin{align}
r_{e_y} &= \exp\left(-\frac{e_y^2}{k_1}\right) \notag \\
r_{e_\psi} = r_{e_\beta} &= 2\exp\left(-\frac{x^2}{k_2}\right) - 1 \label{eq:reward} \\
r_{e_v} &= \exp\left(-\frac{e_v^2}{k_3}\right) \notag
\end{align}

Here, \( f(x) \) represents the common computational form used for both \( r_{e_\psi} \) and \( r_{e_\beta} \), with \( x \) denoting either \( e_\psi \) or \( e_\beta \). This formulation penalizes large angular errors, with a specific structure designed to handle angular wrap-around near \( \pm180^\circ \). The constants \( k_1 \), \( k_2 \) and \( k_3 \) are empirically set to 8, 3200 and 5, respectively.

The total instantaneous reward \( R_{instant} \) is computed as a speed-weighted sum of the partial rewards:
\begin{equation}
\mathcal{R}_{instant} =k_{e_y} r_{e_y} + k_{e_\psi} r_{e_\psi} + k_{e_\beta} r_{e_\beta} +k_{e_v} r_{e_v}
\end{equation}

\begin{equation}
    \mathcal{R}_{instant}^* =
\begin{cases}
    K \cdot \mathcal{R}_{instant},  & |v| < v_{thre}\\
    \mathcal{R}_{instant}, & \text{else}
\end{cases}
\end{equation}

where \( v \) and \( [k_{e_y}, k_{e_\psi}, k_{e_\beta},k_{e_v}] \) denote the vehicle's speed and weight coefficients. This velocity-based scaling encourages the agent not only to follow the desired drift trajectory but also to maintain high speeds—a critical factor in racing performance. To prevent conservative driving behavior, a penalty is imposed if the vehicle speed falls below a threshold velocity $v_{thre}$; in such cases, the total reward is discounted by $K$.

\subsubsection{End Rewards} End rewards are defined as discrete rewards issued at the end of an episode to reflect the overall success or failure of the vehicle's behavior. These rewards serve as high-level guidance signals that help the agent develop a long-term strategy.
Specifically, two types of terminal conditions are considered:

If the vehicle collides with track boundaries or rolls over, a negative reward is assigned. This penalty is inversely proportional to the square of the time duration \( t \) survived within the episode, discouraging early failures and promoting longer, more stable driving. If the vehicle successfully reaches the designated goal or endpoint, it is granted a fixed positive reward. This binary terminal reward structure ensures that the agent is encouraged to complete the full track without crashing, while also learning to avoid dangerous maneuvers that lead to episode termination, as Eq~\eqref{rend}.

\begin{equation}
\mathcal{R}_{end} =
    \begin{cases}
-k_4 \cdot \text{exp}(k_5 t), & collision \ \text{or}\ Rollover\\
k_6 \cdot \text{exp}(k_7\cdot(t-T_{total})), & arrive \ at\ endpoint
\end{cases} \label{rend}
\end{equation}

 The coefficient \( k_7 \) is dependent on the length and characteristics of the racetrack, and it determines the importance of the lap time when the vehicle reaches the goal. The corresponding reward is assigned based on the achieved lap time. Here, \( T_{\text{total}} \) denotes the optimal lap time derived from the MLTDP in Section \ref{sec:planning}.

\textbf{Case:} The set of coefficients \( [k_1, k_2, k_3, k_4, k_5, k_6, k_7] \) shapes the overall reward function. Specifically, \( k_7 \) plays a critical role in evaluating the lap time performance. A higher reward is assigned when the agent's lap time approaches the optimal value defined by the MLTDP.
For instance, on the Lkarting track, the lap time typically ranges from 40 to 50 seconds depending on the driving strategy. Suppose the optimal reference lap time is \( T_{\text{total}} = 46 \) s. If the vehicle completes the episode in 50 s, the final reward is computed as:
\[
R_{{end}} = 500000 \cdot e^{-0.2 \cdot (50 - 46)} = 224664.
\]
This formulation incentivizes the agent to minimize the deviation from the optimal lap time by providing exponentially decreasing rewards for suboptimal performance.

By combining the fine-grained instantaneous rewards with the sparse but critical terminal rewards, the overall reward Eq~\eqref{R} promotes both locally optimal drift control and globally successful racing performance.
\begin{equation}
\mathcal{R} =\mathcal{R}_{instant}^*+ \mathcal{R}_{end} \label{R}
\end{equation}

\subsection{Soft Actor-Critic Training Algorithm}

In this work, we adopt the Soft Actor-Critic (SAC) algorithm~\cite{haarnoja2018soft} to train the continuous drift policy. SAC is an off-policy, entropy-regularized actor-critic method, known for its stability and exploration efficiency in continuous control tasks.
The algorithm optimizes the following objectives:

\paragraph{Critic Loss}  
The Q-function is learned by minimizing the Bellman residual:
\begin{equation}
\mathcal{L}_Q(\theta) = \mathbb{E}_{(s_t, a_t, r_t, s_{t+1}) \sim \mathcal{D}} \left[
    \left( Q_\theta(s_t, a_t) - y_t \right)^2
\right]
\end{equation}
where the target $y_t$ is given by:
\begin{equation}
y_t = r_t + \gamma \, \mathbb{E}_{a_{t+1} \sim \pi_\phi} \left[
    Q_{\bar{\theta}}(s_{t+1}, a_{t+1}) - \alpha \log \pi_\phi(a_{t+1} | s_{t+1})
\right]
\end{equation}
Here, $\bar{\theta}$ denotes the parameters of the target Q-network.
\paragraph{Actor Loss}  
The policy is trained to maximize both expected return and entropy:
\begin{equation}
    \mathcal{L}_\pi(\phi) = \mathbb{E} \left[
        \alpha \log \pi_\phi(a_t|s_t) - Q_\theta(s_t, a_t)
    \right]
\end{equation}

These objectives are optimized iteratively using experience replay and target networks. The SAC framework enables efficient and robust training of the drift control policy in a continuous action space.

\subsection{Track-guided Deep Reinforcement Learning via progressive racetracks}\label{Track_guided}

\begin{figure*}
\centering
\includegraphics[width=\textwidth]{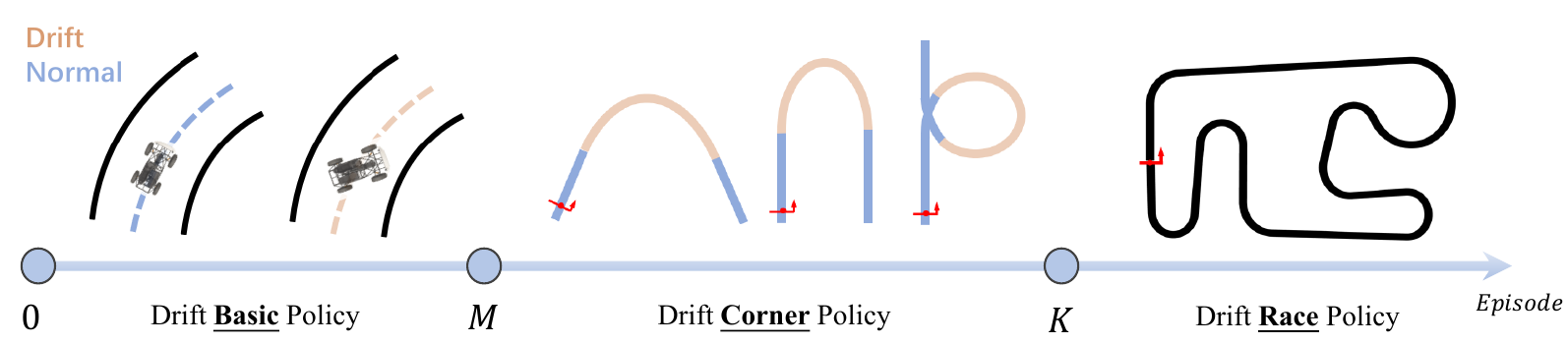}
\caption{TgRL training method. The TgRL training method is designed as a progressive learning framework in which the blue segments of the track represent normal steady-state driving and the orange segments indicate drifting states. The reinforcement learning (RL) agent is trained in a step-by-step manner across three key stages. During the basic drift control policy learning stage (episodes 0 to M), the vehicle primarily learns how to manipulate steering and throttle inputs to induce tire slip and generate a large sideslip angle. In the subsequent drift cornering policy learning stage (episodes M to P), the vehicle focuses on utilizing the drifting posture to navigate corners more effectively, aiming to quickly adjust its heading for optimal corner exit. Finally, in the drift racing policy learning stage (after episode P), the vehicle seeks to minimize lap time by strategically employing drifting maneuvers. In this stage, the agent not only controls the current drifting posture but also anticipates the geometry and timing of upcoming corners to optimize its overall racing performance.}
\label{track_train}
\end{figure*}

In the drift training process, we propose a Track-Guided Reinforcement Learning mechanism inspired by the curriculum \cite{Huang2024} adopted by professional human racing drivers. This mechanism decomposes the learning task into three progressive stages, each associated with a specific type of track and drift skill objective.

In the first stage (from 0 to M episodes), a simplified straight-line track is used to enable the agent to acquire a basic drift control policy. The primary objective at this stage is for the agent to experience and recognize the dynamics of aggressive lateral slip, particularly characterized by large sideslip angles and wheel lock-induced oversteer. This phase helps the agent establish fundamental control over vehicle instability and learn how to recover or maintain drift under simplified conditions.

In the second stage (from M to P episodes), a series of hairpin curves with varying levels of difficulty are introduced. This stage focuses on the development of a drift cornering policy, which teaches the agent how to use controlled drifting to negotiate challenging turns. Through repeated interactions with these geometrically diverse corners, the agent learns when and how to initiate a drift, maintain the slip angle, and execute a smooth corner exit. The control policy developed in this stage incorporates both spatial anticipation and dynamic adaptation, closely resembling human decision-making in cornering maneuvers.

In the third stage (after P episodes), we construct a scaled-down karting-style closed racing circuit to simulate a full racing scenario. The racetrack is built from \textit{Lkarting} park, which can be seen in Fig~\ref{lkarting_tracktest}(a). At this stage, the agent is required to integrate the previously acquired skills and develop a comprehensive race policy that balances high-speed acceleration on straight segments with efficient drift-based cornering. The objective is to complete the race in the shortest possible time while maintaining vehicle stability and trajectory accuracy throughout the course. This stage represents the culmination of the training curriculum and closely approximates real-world autonomous racing tasks.

Fig~\ref{track_train} illustrates the three stages of the proposed track-guided training paradigm. While this track design may appear as a simple geometric progression in terms of difficulty, it is, in fact, a curriculum meticulously structured around the increasing complexity of drift behavior learning. The design mirrors the way human drivers incrementally master drifting—from basic control, to cornering, and ultimately to race-level decision-making—demonstrating how learning can be effectively guided through structured interaction with progressively challenging environments.

\section{Validation Results}\label{sec:validation}

\begin{figure}
\centering
\includegraphics[width=\columnwidth]{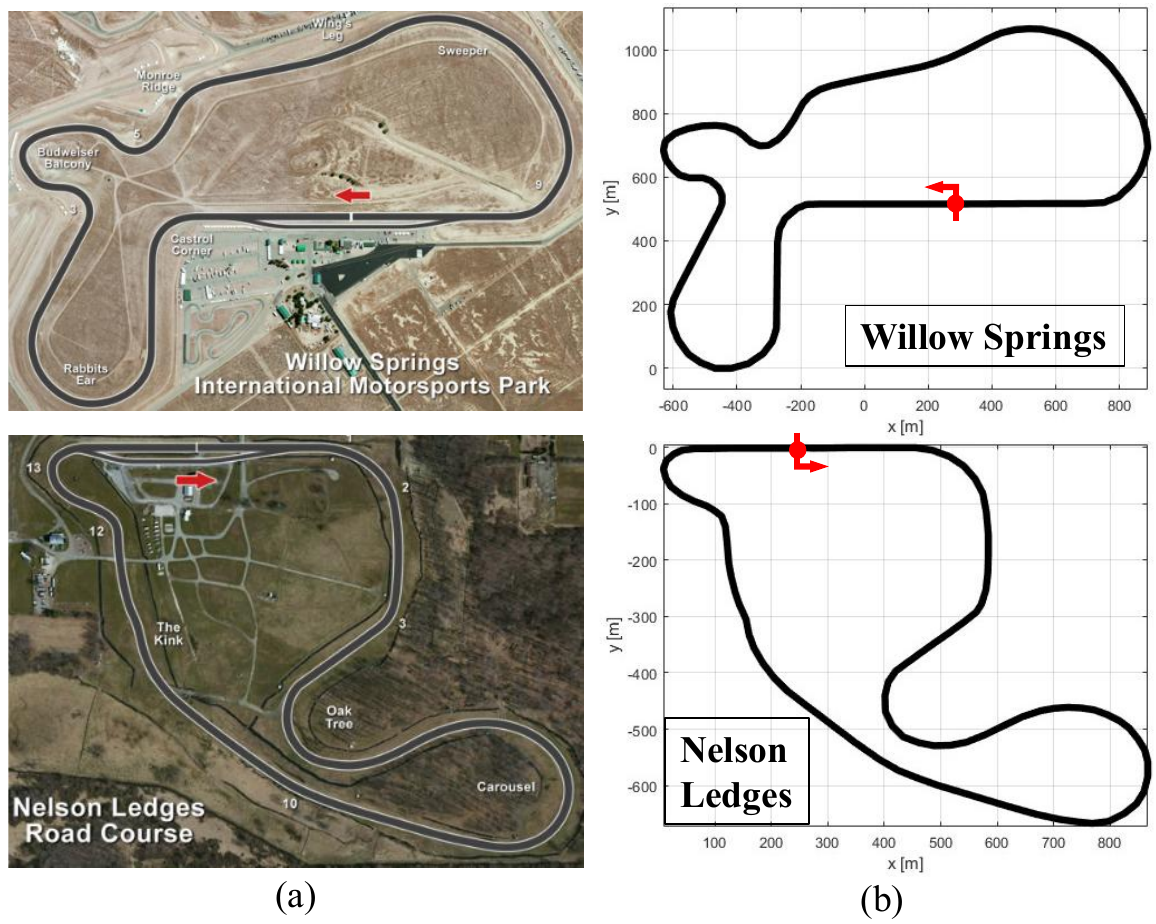}
\caption{Racetracks for test. (a) Satellite image of the racetrack; (b) Simulation track layout.}
\label{tracktest}
\end{figure}

\subsection{Implementation Details}
The MLTDP optimization problem in Sec.~\ref{sec:planning} is solved using SNOPT \cite{snopt}, a solver designed for large-scale nonlinear optimization problems. The MLTDP procedure is implemented in MATLAB, and the resulting solutions are used as reference data for subsequent training. This MLTDP process is based on the drift model established in Sec.~\ref{sec:modelling}.

To validate the proposed training framework, extensive simulation experiments were conducted using the CARLA simulator \cite{carla}, which provides rich APIs for vehicle steering, throttle, braking control, and tire–road friction modeling. The reinforcement learning algorithms interact with CARLA through these interfaces, enabling closed-loop policy training and model weight storage. Both the training and testing processes of the reinforcement learning algorithms are carried out within the CARLA and Python environments, where the built-in vehicle model provided by CARLA is adopted.

Custom track environments, including Lkarting, Willow Springs, and Nelson Ledges, were designed in RoadRunner and subsequently imported into the CARLA simulation platform for evaluation, as shown in Fig. \ref{tracktest}.

Each simulation episode terminates when the agent either reaches the designated endpoint, encounters a collision, or exhausts the predefined maximum number of steps. Detailed RL hyperparameter configurations are also listed in Table~\ref{canshu}. In particular, the actor–critic update ratio is set to 1:5, meaning that the actor network is updated once every five critic updates. This design allows the critic to better approximate the value function before each policy update, thereby improving training stability and mitigating the modeling uncertainties and mismatch between CARLA dynamics and drift dynamics in Sec. \ref{sec:modelling}. All key parameter values used in Markov Decision Process are also summarized in Table~\ref{canshu}.

\subsection{Equipment Configurations}
All algorithms were trained and evaluated on a workstation equipped with an Intel Core i9-9900K CPU, 16 GB of RAM, and an NVIDIA GTX 2080 graphics card. 

\begin{figure}
\centering
\includegraphics[width=\columnwidth]{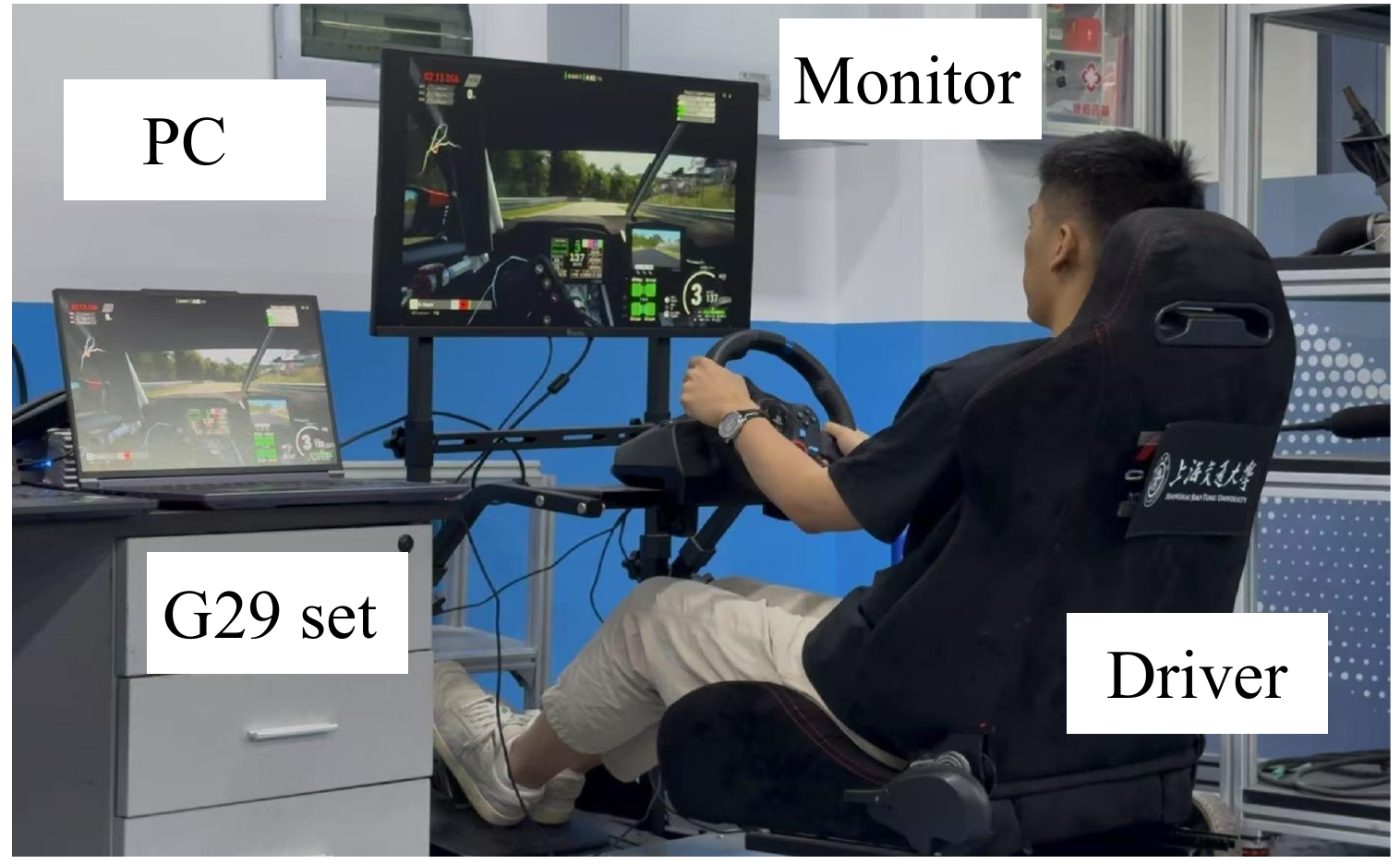}
\caption{A human driver operating a Logitech G29 simulator setup to perform tests on different tracks.}
\label{fig:drvier}
\end{figure}

We collected human driving performance data under the same simulation environment to enable a more comprehensive comparison with the proposed drifting controller. The driver data were acquired using a simulator equipped with a Logitech G27 control setup, as shown in Fig.~\ref{fig:drvier}.

\begin{table}[t!]
\caption{\textbf{Hyperparameter Settings for RL Training.}} 
\centering
\resizebox{\columnwidth}{!}{%
\begin{tabular}{@{} l c @{}} 
\toprule
\textbf{Parameter} & \textbf{Value} \\
\midrule
$\lambda_{\tau,t}$ in ASS & 0.2 \\
$[v_{\max}, v_{\min}]$ in ASS & [16.67, 20] km/h \\
State error reward weights $[k_1, k_2, k_3]$ & [8, 3200, 5] \\
Instant reward weights $[k_{e_y}, k_{e_\psi}, k_{e_\beta}, k_{e_v}]$ & [20, 40, 20, 40] \\
Velocity discounted weight $K$ & 0.5 \\
Velocity threshold $v_{\text{thre}}$ & 5 m/s \\
End reward weights $[k_4, k_5, k_6, k_7]$ & [$10^5$, -0.05, $10^5$, -0.2]\\
Minibatch size & 128 \\
Replay buffer size & 300000 \\
Discount factor & 0.995 \\
Actor : Critic update ratio & 1:5 \\
Target network update parameter & 0.005 \\
Actor learning rate & 0.0001 \\
Critic learning rate & 0.001 \\
\bottomrule
\end{tabular}%
}
\label{canshu}
\end{table}

\subsection{Lkarting Result Analysis}

\begin{figure*}
\centering
\includegraphics[width=\textwidth]{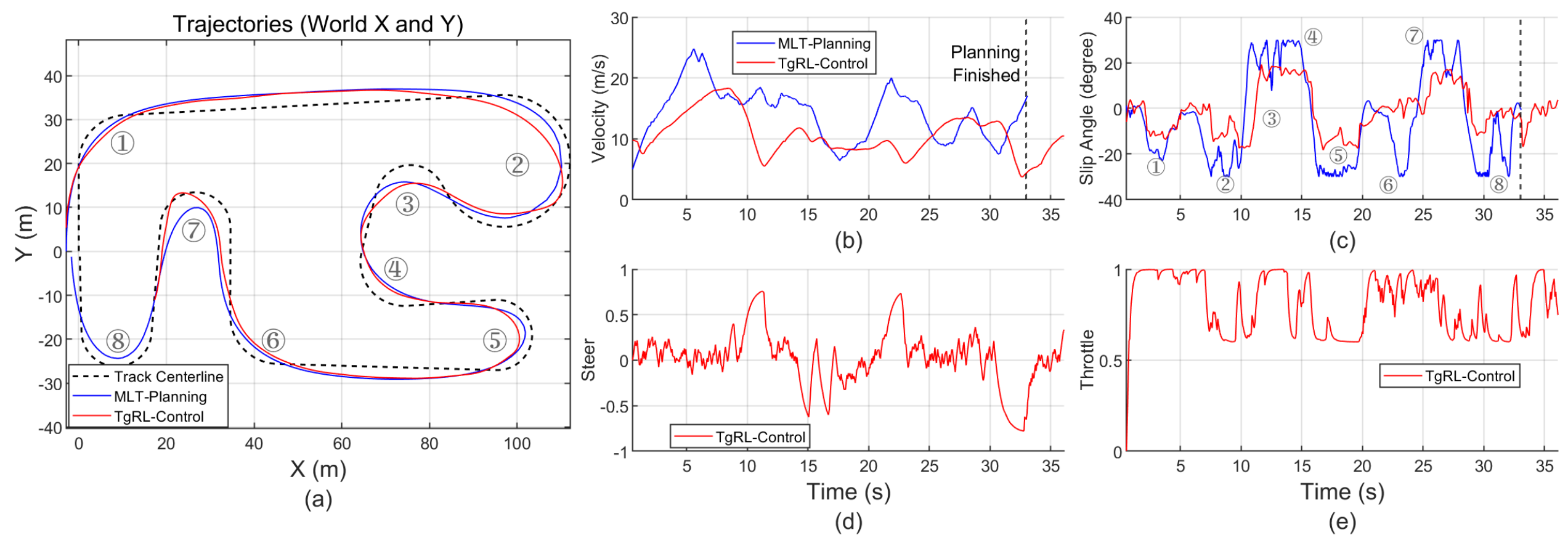}
\caption{Drift test on Lkarting racetrack. (a) Trajectories; (b) Velocity; (c) Sideslip angle; (d) Steer; (e) Throttle.}
\label{lkarting_tracktest}
\end{figure*}

The core concept of the MLT-TgRL drift control framework is to employ a model-based optimization method to guide reinforcement learning in acquiring drifting strategies. On the small-scale racetrack, \textit{Lkarting}, the experimental results, including control inputs (throttle and steering) and key state variables (velocity and sideslip angle) shown in Fig~\ref{lkarting_tracktest}, validate the effectiveness of the proposed framework.

\begin{figure}
\centering
\includegraphics[width=\columnwidth]{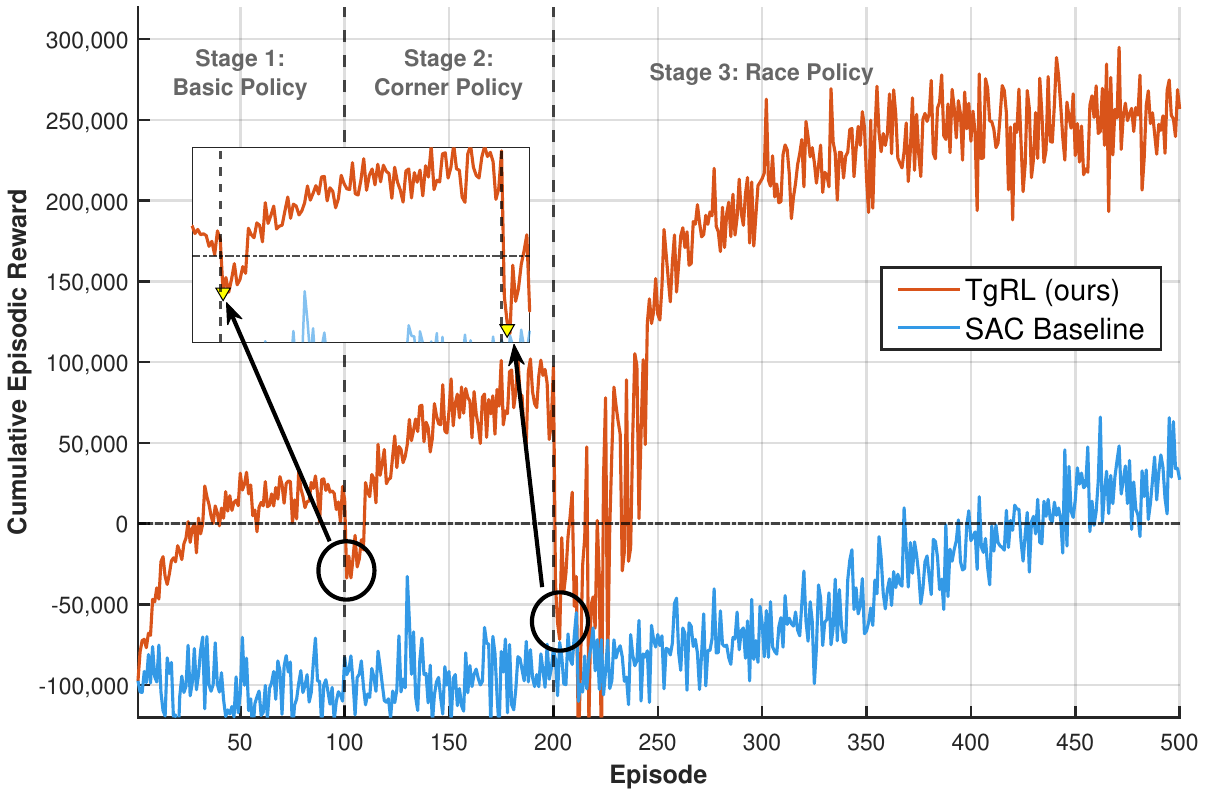}
\caption{Training convergence of the proposed TgRL compared to the SAC baseline. The training process of TgRL is divided into three progressive curriculum stages. The inset plot details the transfer shocks (yellow triangles) that occur at the boundaries of tasks with increasing complexity (Episodes 100 and 200).}
\label{RL_line}
\end{figure}

As illustrated in Fig. \ref{RL_line}, the proposed TgRL significantly outperforms the standard SAC baseline in both learning efficiency and final asymptotic performance. Due to the strict collision penalties within the complex racing environment, the standard SAC agent struggles with early-stage exploration, remaining trapped in a suboptimal negative reward space for over 350 episodes.
In contrast, TgRL leverages a progressive curriculum to build fundamental skills efficiently. While the transition to more complex tasks (from Basic to Corner, and Corner to Race policy) introduces temporary performance drops, or transfer shocks, at episodes 100 and 200 (highlighted in the inset), the agent rapidly adapts to the new dynamics. Ultimately, TgRL overcomes these bottlenecks and converges to a highly optimal race policy, demonstrating the necessity and effectiveness of curriculum-based learning for autonomous racing tasks.

For detailed evaluation of planning and control performance, the \textit{Lkarting} track is divided into eight corners (\ding{192}-\ding{199}), as illustrated in Fig~\ref{lkarting_tracktest}(a). The trajectory generated by MLT-Planning is relatively smooth because it is obtained by solving an optimal control problem. In contrast, the TgRL-Control trajectory demonstrates actual drifting behavior executed by the RL agent. As seen in corners \ding{192}, \ding{194}, \ding{195}, \ding{196}, \ding{197}, and \ding{199}, the TgRL drift controller achieves precise drifting comparable to MLT-Planning. However, larger discrepancies are observed in corners \ding{193} and \ding{198}, where Fig~\ref{lkarting_tracktest}(c) shows that insufficient tracking of the desired sideslip angle prevents the vehicle from maintaining an optimal drifting state (i.e., large sideslip angles) through these curves. Moreover, the TgRL controller does not explicitly anticipate curvature changes derived from upcoming waypoints, which may explain its delayed initiation of drifting before sharp turns, thus reducing cornering efficiency. 

From Figs.~\ref{lkarting_tracktest}~(b) and (c), MLT-Planning achieves a smoother trajectory, higher velocities, and larger sideslip angles, while TgRL-Control exhibits more conservative dynamics (e.g., a maximum sideslip angle of approximately $18^\circ$). This discrepancy is likely due to differences in the vehicle models used for planning and control. Nevertheless, the TgRL agent successfully learns an effective drifting policy. Notably, the planned lap time is approximately $33~\mathrm{s}$, while TgRL-Control completes the lap in about $36.5~\mathrm{s}$.

\begin{figure*}
\centering
\includegraphics[width=\textwidth]{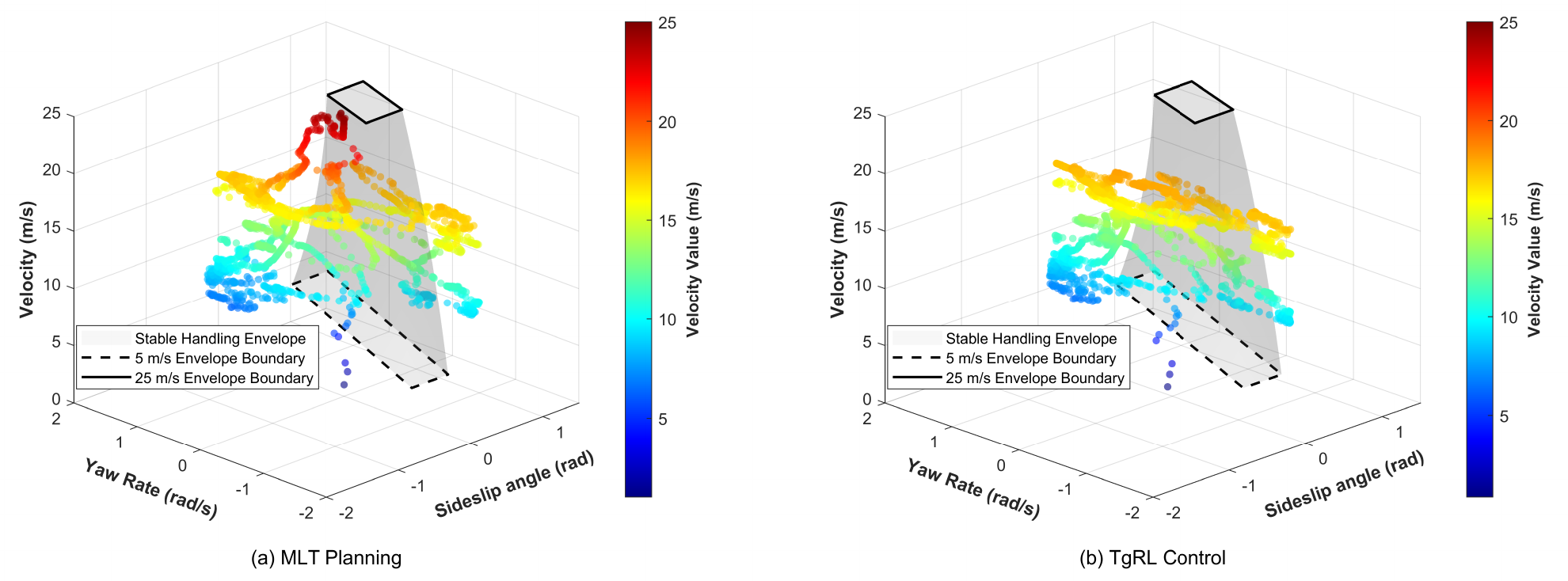}
\caption{Three-dimensional phase diagram of vehicle dynamics states.}
\label{fig:phase_plane_3d}
\end{figure*}

Figure \ref{fig:phase_plane_3d} details the vehicle dynamics in a state space defined by velocity, sideslip angle, and yaw rate. The shaded area represents the stable handling envelope. During regular driving, the vehicle operates exclusively within this steady state region to maintain tire grip and overall stability. However, the plotted trajectories reveal that both the MLT planning and TgRL control methods frequently drive the vehicle outside this predefined boundary. Operating extensively in this nonlinear external region with elevated sideslip angles is the defining feature of a drift. The data distribution clearly proves that the vehicle exceeds normal handling limits and is actively maintained in a continuous drift state rather than executing standard aggressive cornering.



\subsection{Different Racetrack Result Analysis}

To further validate the effectiveness and generalization capability of the proposed drift control framework, additional experiments were conducted on two well-known circuits, Willow Springs and Nelson Ledges, in addition to the previously tested \textit{Lkarting} track. The layouts of these racetracks are shown in Fig~\ref{tracktest}. We evaluate three configurations: MLT (Planning), which represents a trajectory generated offline by solving an optimal control problem; MLT+TgRL (Planning+Control), our proposed approach that integrates model-based trajectory optimization with a reinforcement learning controller for real-time execution; and two baselines, including a reinforcement learning drifting controller proposed by Cai \textit{et al.}~\cite{cai2020} and the \textit{Human Driver} data collected from a professional driver. A summary of all performance metrics is provided in Table~\ref{resultsTable}, where we report lap times, maximum and average sideslip angles, peak and mean vehicle speeds, yaw angle tracking errors, and lateral offset relative to the track centerline, with the best performance highlighted in bold.

The results indicate that the proposed MLT+TgRL controller consistently achieves superior lap times compared to both learning-based baselines, while approaching the theoretical lower bound set by MLT Planning, which does not account for execution uncertainty or actuation constraints. On \textit{Lkarting}, the MLT+TgRL framework completes laps in an average of $55.4~\mathrm{s}$, outperforming Cai\_DRIFT by approximately $10\%$, while achieving a comparable average sideslip angle of $15.9^\circ$, close to the optimal planning reference of $29.8^\circ$. Similar improvements are observed on Willow Springs and Nelson Ledges, where our controller reduces minimum lap time by $6\%$ and $7\%$, respectively, compared with the Cai\_DRIFT baseline, while maintaining higher average sideslip angles than the human driver policy. Although Cai\_DRIFT occasionally reached higher speeds, its inability to maintain stable yaw tracking resulted in larger errors and wider lateral deviations, underscoring the advantage of integrating model-based trajectory optimization with RL-based control execution. The proposed controller demonstrates consistent path-tracking accuracy, with yaw angle error averaging $6.2^\circ$ on \textit{Lkarting}, $8.9^\circ$ on \textit{Willow Springs} and $5.1^\circ$ on \textit{Nelson Ledges}, and minimal lateral offset fluctuations, further confirming its stability and precision.

\begin{table*}[!htbp]
\centering 
\footnotesize 
\caption{\textbf{The racetrack testing results of different methods. The sideslip angle Avg (Laps) means the average of the max sideslip angle during ten testing laps.}}
\label{resultsTable}
\begin{tabular*}{\textwidth}{@{\extracolsep{\fill}} cccccc cccccc @{}}
\toprule
\multirow{2}{*}[-0.1cm]{Track} & \multirow{2}{*}[-0.1cm]{Method}
& \multicolumn{2}{c}{Lap Time (s)}  
& \multicolumn{2}{c}{Sideslip Angle $(^\circ)$} 
& \multicolumn{2}{c}{Speed (m/s)} 
& \multicolumn{2}{c}{Yaw Angle Error $(^\circ)$} 
& \multicolumn{2}{c}{L.O. to CL (m)} \\
\cmidrule(r){3-4} \cmidrule(r){5-6} \cmidrule(r){7-8} \cmidrule(r){9-10} \cmidrule(r){11-12}
&& Min & Avg & Max & Avg (Laps) & Max & Avg & Max & Avg & Max & Avg \\
\midrule
\multirow{4}{*}{Lkarting} 
    & MLT Planning      & 33.1 & - & 29.9 & 29.8   & 25.0 & 13.1 & 0 & 0  & 3.0 & 2.7 \\
    & \textbf{MLT+TgRL (Ours)}   & \textbf{37.2} & \textbf{45.4} & \textbf{23.5} & 15.9 & \textbf{33.8} & \textbf{9.7} & \textbf{10.0} & \textbf{6.2} & \textbf{+0.1} & \textbf{-0.2} \\
    & Human Driver             & 51.3 & 54.2 & 14.6 & 10.4 & 24.3 & 8.0 & 22.4 & 12.1 & -0.7 & -1.0 \\
    & Cai\_DRIFT \cite{cai2020} & 47.6 & 51.2 & 17.8 & \textbf{16.3} & 28.6 & 8.4 & 12.8 & 7.2 & -0.2 & -0.6 \\
\midrule

\multirow{4}{*}{Willow Springs} 
    & MLT Planning             & 155 & - & 29.9 & -   & 35.0 & 20.1 & 0 & 0   & 2.7 & 1.9 \\
    & \textbf{MLT+TgRL (Ours)}   & \textbf{181.1} & \textbf{197.5} & \textbf{18.2} & \textbf{17.2} & 38.8 & 17.6 & 13.8 & \textbf{8.9} & \textbf{+0.5} & \textbf{+0.2} \\
    & Human Driver               & 202.9 & 210.0 & 11.1 & 6.4 & 33.4 & 15.6 & 21.1 & 14.2 & +1.2 & -0.2 \\
    & Cai\_DRIFT \cite{cai2020} & 193.1 & 197.7 & 15.2 & 14.3 & \textbf{42.6} & \textbf{17.9} & \textbf{11.1} & 9.2 & +0.6 & +0.3 \\
\midrule

\multirow{4}{*}{Nelson Ledges} 
    & MLT Planning             & 142 & - & 29.9 & -  & 35.0 & 25.4 & 0 & 0   & 3.1 & 2.1 \\
    & \textbf{MLT+TgRL (Ours)}   & \textbf{158.8} & \textbf{169.2} & \textbf{19.1} & \textbf{18.2} & \textbf{44.0} & 22.1 & \textbf{9.6} & \textbf{5.1} & -0.5 & \textbf{-0.7} \\
    & Human Driver               & 180.2 & 196.5 & 9.2 & 5.6 & 39.8 & 18.7 & 17.1 & 10.2 & -1.5 & -2.0 \\
    & Cai\_DRIFT \cite{cai2020} & 171.1 & 174.5 & 14.0 & 11.8 & 41.0 & \textbf{24.4} & 11.1 & 7.0 & \textbf{+0.1} & -1.1 \\
\bottomrule
\end{tabular*}
\end{table*}

\begin{figure*}
\centering
\includegraphics[width=\textwidth]{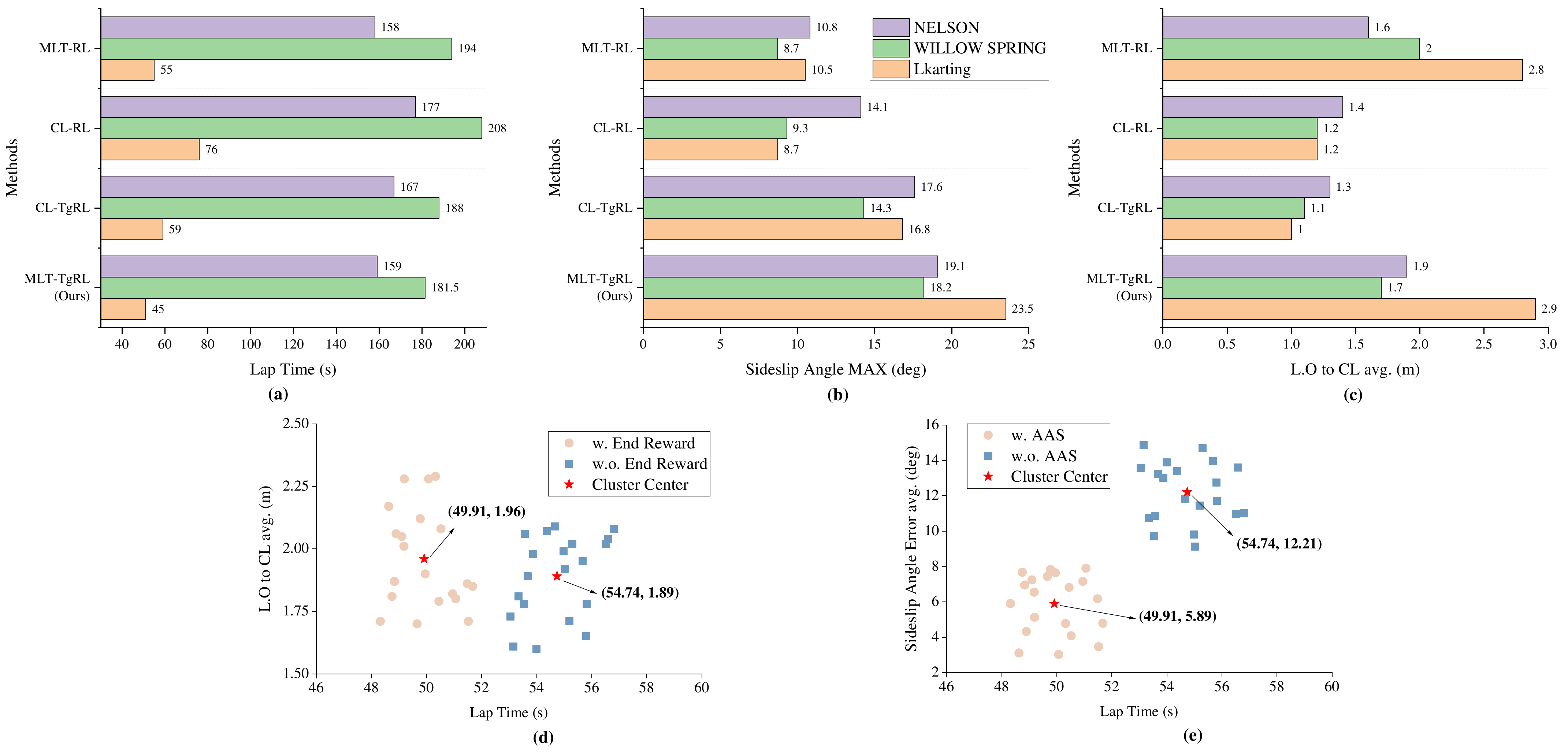}
\caption{The ablation experiment of our methods: (a) Lap time; (b) Sideslip angle (max); (c) Lateral offset to centerline (avg). (d) Impact of End reward to the lap time and L.O; (e) Impact of AAS to the lap time and sideslip angle.}
\label{ablation1}
\end{figure*}

Taken together, these findings demonstrate that the MLT+TgRL framework successfully bridges the gap between theoretically optimal planning and practical drift execution. By leveraging the model-based planner to provide dynamically feasible references and the RL controller to adapt to unmodeled dynamics and actuation delays, the proposed method achieves aggressive drift maneuvers with improved stability and repeatability. Its superior performance across three racetracks of varying geometry and curvature density highlights the generalization capability of this approach and its potential for real-world autonomous racing applications.

\subsection{Ablation Study}

To further investigate the contribution of individual components in the proposed framework, we performed a comprehensive ablation study by systematically disabling or modifying specific modules and comparing the resulting performance. The comparison includes four configurations: \textbf{MLT+TgRL (Proposed)}, which combines model-based trajectory generation with our multiple-stage track-guided reinforcement learning controller; \textbf{CL+TgRL}, in which the RL agent is trained to follow the track centerline instead of the optimized drift trajectory; \textbf{CL+RL}, which replaces the TgRL controller with a conventional RL agent trained on centerline references; and \textbf{MLT+RL}, where the optimal MLT trajectory is paired with a standard RL controller. Fig.~\ref{ablation1} summarizes the results across three racetracks: \textit{Lkarting}, \textit{Nelson Ledges}, and \textit{Willow Springs}.

The lap time comparison in Fig~\ref{ablation1}(a) clearly illustrates the advantage of the proposed MLT+TgRL framework. Agents trained with MLT-based guidance, regardless of whether TgRL or standard RL is used, consistently outperform those trained with centerline references, highlighting the critical role of model-based trajectory optimization in shaping efficient and dynamically feasible drift paths. Among these, TgRL further improves lap time by leveraging its stage-wise, track-guided training strategy, enabling better adaptation to sharp curvature transitions and effective exploitation of vehicle dynamics near the handling limit. By contrast, agents trained solely on centerline references (CL+RL) fail to capture drift-specific strategies, instead resorting to conservative cornering behaviors that result in significantly slower laps.

The sideslip angle analysis in Fig.~\ref{ablation1}(b) supports this observation. The proposed framework achieves the highest maximum sideslip angles, demonstrating its capability to sustain aggressive drift maneuvers while maintaining vehicle stability. In contrast, CL-based agents exhibit notably smaller sideslip angles, confirming that a lack of drift trajectory guidance limits their ability to perform high-angle drifts through tight corners. Figure~\ref{ablation1}(c) further highlights this effect, where the average lateral offset to the track centerline is minimized when agents follow MLT-generated references, indicating a closer adherence to dynamically optimal paths and improved precision in cornering execution.

Figures~\ref{ablation1}(d) and \ref{ablation1}(e) provide deeper insights into the effectiveness of the proposed reward shaping strategies. In Fig~\ref{ablation1}(d), we compare agents trained with and without the End reward over 20 evaluation laps. The inclusion of the End reward significantly improves lap time consistency and reduces lateral deviation, suggesting that terminal performance cues are essential for encouraging globally optimal trajectory adherence. Similarly, Fig~\ref{ablation1}(e) shows that incorporating the Adaptive Angle Shaping (AAS) reward leads to reduced sideslip angle errors, allowing the controller to sustain high-angle drifts with improved accuracy. These enhancements collectively contribute to shorter lap times and greater overall stability.

Overall, the ablation study confirms that each component of the proposed MLT+TgRL framework plays a critical role in achieving high-performance drift control. The model-based trajectory generator provides physically consistent and strategically meaningful guidance, while the track-guided RL architecture enables the controller to refine complex drift strategies through staged learning. Reward shaping techniques, such as the End reward and AAS, further accelerate convergence and improve robustness, resulting in superior performance across racetracks of varying complexity. These findings validate the necessity of integrating trajectory optimization, tailored reinforcement learning design, and carefully engineered rewards to achieve reliable, high-speed autonomous drifting.

\section{Conclusion}\label{sec:conclusion}
This paper proposes a vehicle drifting control framework that integrates a physics-based optimization method with a reinforcement learning (RL) controller, referred to as MLT-TgRL. The Minimum Lap Time (MLT) module is designed to optimize racing trajectories under large sideslip angle conditions and to provide prior knowledge for training the RL-based drifting controller. The RL controller outputs steering and throttle commands based on vehicle state information. Its key advantage lies in the capability of neural networks to process high-dimensional state inputs and directly generate low-dimensional control outputs, making it well-suited for nonlinear vehicle control tasks.

In addition, a track-guided reinforcement learning training strategy is proposed to enable the agent to gradually acquire drifting skills. This guided training approach helps the agent better understand the application of drifting maneuvers in race-like scenarios. Such a strategy often enhances the vehicle’s ability to maintain larger sideslip angles while cornering, leading to reduced lap times.

The contribution of this work is twofold: it introduces a novel control framework for autonomous drifting and presents a training paradigm tailored for RL-based vehicle controllers. This lays a practical foundation for using reinforcement learning in real-world vehicle control systems.

Nevertheless, the current approach has limitations. Future work will focus on improving the generalization capability of the controller across different vehicle models, varying road surface conditions, and diverse tire characteristics. Moreover, we plan to implement real-time deployment through hardware-in-the-loop (HIL) experiments.
\printcredits
\bibliographystyle{model1-num-names}

\bibliography{ref}

@ARTICLE{Li_2024_multiobjective,
	author = {Li, Jie and Wu, Xiaodong and Xu, Min and Liu, Yonggang},
	title = {Multiobjective Eco-Driving Strategy for Connected and Automated Electric Vehicles Considering Complex Urban Traffic Influence Factors},
	year = {2024},
	journal = {IEEE Transactions on Transportation Electrification},
	volume = {10},
	number = {4},
	pages = {10043 – 10058},
	doi = {10.1109/TTE.2023.3349025},
	url = {https://www.scopus.com/inward/record.uri?eid=2-s2.0-85181562163&doi=10.1109%2fTTE.2023.3349025&partnerID=40&md5=13ed9f895bd65d5da4838532ca46bf35},
	type = {Article},
	publication_stage = {Final},
	source = {Scopus},
	note = {Cited by: 5}
}

@ARTICLE{Li_2023_overcoming,
	author = {Li, Jie and Wu, Xiaodong and Fan, Jiawei and Liu, Yonggang and Xu, Min},
	title = {Overcoming driving challenges in complex urban traffic: A multi-objective eco-driving strategy via safety model based reinforcement learning},
	year = {2023},
	journal = {Energy},
	volume = {284},
	doi = {10.1016/j.energy.2023.128517},
	url = {https://www.scopus.com/inward/record.uri?eid=2-s2.0-85166514858&doi=10.1016%2fj.energy.2023.128517&partnerID=40&md5=ddcec4a99c7f7563bb4ad2942c01a487},
	type = {Article},
	publication_stage = {Final},
	source = {Scopus},
	note = {Cited by: 23}
}

@ARTICLE{Li_2022_deep,
	author = {Li, Jie and Wu, Xiaodong and Xu, Min and Liu, Yonggang},
	title = {Deep reinforcement learning and reward shaping based eco-driving control for automated HEVs among signalized intersections},
	year = {2022},
	journal = {Energy},
	volume = {251},
	doi = {10.1016/j.energy.2022.123924},
	url = {https://www.scopus.com/inward/record.uri?eid=2-s2.0-85129074799&doi=10.1016%2fj.energy.2022.123924&partnerID=40&md5=6851ac6dfe0baffbdb87ad5333aefe6a},
	type = {Article},
	publication_stage = {Final},
	source = {Scopus},
	note = {Cited by: 82}
}

@ARTICLE{Li_2023_deep,
	author = {Li, Jie and Fotouhi, Abbas and Pan, Wenjun and Liu, Yonggang and Zhang, Yuanjian and Chen, Zheng},
	title = {Deep reinforcement learning-based eco-driving control for connected electric vehicles at signalized intersections considering traffic uncertainties},
	year = {2023},
	journal = {Energy},
	volume = {279},
	doi = {10.1016/j.energy.2023.128139},
	url = {https://www.scopus.com/inward/record.uri?eid=2-s2.0-85162056297&doi=10.1016%2fj.energy.2023.128139&partnerID=40&md5=166c42d93d01b536fb3ead51b8abdb2c},
	type = {Article},
	publication_stage = {Final},
	source = {Scopus},
	note = {Cited by: 42; All Open Access, Green Open Access}
}

@ARTICLE{Ni2017,
  author={Ni, Jun and Hu, Jibin and Xiang, Changle},
  journal={IEEE Trans. Veh. Technol.}, 
  title={Envelope Control for Four-Wheel Independently Actuated Autonomous Ground Vehicle Through AFS/DYC Integrated Control}, 
  year={2017},
  volume={66},
  number={11},
  pages={9712-9726},
  doi={10.1109/TVT.2017.2723418}}

@ARTICLE{huang2020,
  author={Huang, Yiwen and Chen, Yan},
  journal={IEEE Trans. Veh. Technol.}, 
  title={Vehicle Lateral Stability Control Based on Shiftable Stability Regions and Dynamic Margins}, 
  year={2020},
  volume={69},
  number={12},
  pages={14727-14738},
  doi={10.1109/TVT.2020.3036780}}

@ARTICLE{cheng2025,
  author={Cheng, Shuo and Hu, Bin-Bin and Wei, Heng-Lai and Li, Liang and Lv, Chen},
  journal={IEEE Trans. Veh. Technol.}, 
  title={Deep Learning-Based Hybrid Dynamic Modeling and Improved Handling Stability Assessment for Autonomous Vehicles at Driving Limits}, 
  year={2025},
  volume={74},
  number={4},
  pages={5582-5593},
  doi={10.1109/TVT.2024.3515209}}

@ARTICLE{wu2023,
  author={Wu, Xiaodong and Li, Jie and Su, Chengrui and Fan, Jiawei and Xu, Min},
  journal={IEEE Trans. Veh. Technol.}, 
  title={A Deep Reinforcement Learning Based Hierarchical Eco-Driving Strategy for Connected and Automated HEVs}, 
  year={2023},
  volume={72},
  number={11},
  pages={13901-13916},
  doi={10.1109/TVT.2023.3283617}}

@ARTICLE{ma2024,
  author={Ma, Qingwen and Yin, Xin and Zhang, Xinglong and Xu, Xin and Yao, Xinxin},
  journal={IEEE Trans. Veh. Technol.}, 
  title={Game-Theoretic Receding-Horizon Reinforcement Learning for Lateral Control of Autonomous Vehicles}, 
  year={2024},
  volume={73},
  number={10},
  pages={14547-14562},
  doi={10.1109/TVT.2024.3412530}}

@ARTICLE{Shan2024,
  author={Shan, Zitong and Zhao, Jian and Zhu, Bing and Lv, Chen and Zhao, Yang and Ge, Linhe and Zhong, Shouren},
  journal={IEEE Trans. Veh. Technol.}, 
  title={Safe and Efficient Trajectory Planning Considering Longitudinal and Lateral Coupled Limits}, 
  year={2024},
  volume={73},
  number={7},
  pages={10714-10719},
  doi={10.1109/TVT.2024.3367125}}

@article{snopt,
author = {Gill, Philip E. and Murray, Walter and Saunders, Michael A.},
title = {SNOPT: An SQP Algorithm for Large-Scale Constrained Optimization},
journal = {SIAM Rev.},
volume = {47},
number = {1},
pages = {99-131},
year = {2005},
doi = {10.1137/S0036144504446096},
URL = { 
        https://doi.org/10.1137/S0036144504446096
},
eprint = { 
        https://doi.org/10.1137/S0036144504446096
}
}

@ARTICLE{Li2025,
  author={Li, Jie and Wu, Xiaodong and Bai, Xianxu and Liu, Yonggang and Xu, Min},
  journal={IEEE Trans. Intell. Transp. Syst.}, 
  title={Intelligent Eco-Driving Control for Urban CAVs Using a Model-Based Controller Assisted Deep Reinforcement Learning}, 
  year={2025},
  volume={26},
  number={6},
  pages={7624-7639},
  doi={10.1109/TITS.2025.3559916}}

@InProceedings{carla,
  title = 	 {{CARLA}: {An} Open Urban Driving Simulator},
  author = 	 {Dosovitskiy, Alexey and Ros, German and Codevilla, Felipe and Lopez, Antonio and Koltun, Vladlen},
  booktitle = 	 {Proc.s of the 1st Annu. Conf. on Robot Learn.},
  pages = 	 {1--16},
  year = 	 {2017},
  editor = 	 {Levine, Sergey and Vanhoucke, Vincent and Goldberg, Ken},
  volume = 	 {78},
  series = 	 {Proceedings of Mach. Learn. Res.},
  month = 	 {13--15 Nov},
  publisher =    {PMLR},
}

@InProceedings{haarnoja2018soft,
  title = 	 {Soft Actor-Critic: Off-Policy Maximum Entropy Deep Reinforcement Learning with a Stochastic Actor},
  author =       {Haarnoja, Tuomas and Zhou, Aurick and Abbeel, Pieter and Levine, Sergey},
  booktitle = 	 {Proc. of the 35th Int. Conf. on Mach. Learn.},
  pages = 	 {1861--1870},
  year = 	 {2018},
  editor = 	 {Dy, Jennifer and Krause, Andreas},
  volume = 	 {80},
  series = 	 {Proc. of Mach. Learn. Res.},
  month = 	 {10--15 Jul},
  publisher =    {PMLR},
}

@ARTICLE{Huang2024,
  author={Huang, Wenhui and Liu, Haochen and Huang, Zhiyu and Lv, Chen},
  journal={IEEE Trans. Intell. Transp. Syst.}, 
  title={Safety-Aware Human-in-the-Loop Reinforcement Learning With Shared Control for Autonomous Driving}, 
  year={2024},
  volume={25},
  number={11},
  pages={16181-16192},
  doi={10.1109/TITS.2024.3420959}}

@ARTICLE{Weber2024,
  author={Weber, Trey P. and Aggarwal, Rajan K. and Gerdes, J. Christian},
  journal={IEEE Trans. Intell. Veh.}, 
  title={Human-Inspired Autonomous Racing in Low Friction Environments}, 
  year={2024},
  volume={},
  number={},
  pages={1-14},
  doi={10.1109/TIV.2024.3462253}}

@ARTICLE{Kabzan2019,
  author={Kabzan, Juraj and Hewing, Lukas and Liniger, Alexander and Zeilinger, Melanie N.},
  journal={IEEE Robot. and Automat. Lett.}, 
  title={Learning-Based Model Predictive Control for Autonomous Racing}, 
  year={2019},
  volume={4},
  number={4},
  pages={3363-3370},
  doi={10.1109/LRA.2019.2926677}}

@article{Lu2025, 
  author = {Lu, Hangyu and Wu, Xiaodong and Zhao, Sheng and Yan, Liang and Lu, Jianwei},
  title = {Controlling nonlinear vehicular motions by exploiting linearized feedback law under delay-tolerance: stability, gain-scheduling, and validation},
  journal = {Meccanica},
  year = {2025},
  volume = {},
  number = {},
  pages = {},
  url = {https://doi.org/10.1007/s11012-025-02027-w}
}

@article{toth2024,
  author = {Tóth, Sándor Hunor and Viharos, Zsolt József and Bárdos, Ákos and Szalay, Zsolt},
  title = {Sim-to-Real Application of Reinforcement Learning Agents for Autonomous, Real Vehicle Drifting},
  journal = {Vehicles},
  year = {2024},
  volume = {6},
  number = {2},
  pages = {781--798},
  doi = {10.3390/vehicles6020037}
}

@ARTICLE{cai2020,
  author={Cai, Peide and Mei, Xiaodong and Tai, Lei and Sun, Yuxiang and Liu, Ming},
  journal={IEEE Robot. and Automat. Lett.}, 
  title={High-Speed Autonomous Drifting With Deep Reinforcement Learning}, 
  year={2020},
  volume={5},
  number={2},
  pages={1247-1254},
  doi={10.1109/LRA.2020.2967299}}

@article{HOU2022,
title = {Autonomous driving at the handling limit using residual reinforcement learning},
journal = {Adv. Eng. Inform.},
volume = {54},
pages = {101754},
year = {2022},
issn = {1474-0346},
doi = {https://doi.org/10.1016/j.aei.2022.101754},
url = {https://www.sciencedirect.com/science/article/pii/S1474034622002129},
author = {Xiaohui Hou and Junzhi Zhang and Chengkun He and Yuan Ji and Junfeng Zhang and Jinheng Han}
}

@article{LI2023,
title = {Planning and control of drifting-based collision avoidance strategy under emergency driving conditions},
journal = {Control Eng. Pract.},
volume = {139},
pages = {105625},
year = {2023},
issn = {0967-0661},
doi = {https://doi.org/10.1016/j.conengprac.2023.105625},
url = {https://www.sciencedirect.com/science/article/pii/S0967066123001946},
author = {Daofei Li and Jiajie Zhang and Siyuan Lin}
}

@ARTICLE{zhao2024_Autonomous,
  author={Zhao, Xuanming and Chen, Guoying and Gao, Zhenhai and Yao, Jun and Gao, Zheng and Hua, Min},
  journal={IEEE Trans. on Transp. Electrific.}, 
  title={Autonomous Obstacle Avoidance for Distributed Drive Electric Vehicles via Dynamic Drifting}, 
  year={2024},
  volume={10},
  number={4},
  pages={8893-8906},
  doi={10.1109/TTE.2024.3380629}}

@article{AJANOVIC2023,
title = {Search-based task and motion planning for hybrid systems: Agile autonomous vehicles},
journal = {Eng. Appl. of Artif. Intell.},
volume = {121},
pages = {105893},
year = {2023},
issn = {0952-1976},
doi = {https://doi.org/10.1016/j.engappai.2023.105893},
url = {https://www.sciencedirect.com/science/article/pii/S0952197623000775},
author = {Zlatan Ajanović and Enrico Regolin and Barys Shyrokau and Hana Ćatić and Martin Horn and Antonella Ferrara}
}

@article{Hindiyeh2014,
    author = {Hindiyeh, Rami Y. and Christian Gerdes, J.},
    title = {A Controller Framework for Autonomous Drifting: Design, Stability, and Experimental Validation},
    journal = {J. of Dyn. Syst., Meas., and Control},
    volume = {136},
    number = {5},
    pages = {051015},
    year = {2014},
    month = {07},
    issn = {0022-0434},
    doi = {10.1115/1.4027471},
    url = {https://doi.org/10.1115/1.4027471},
    eprint = {https://asmedigitalcollection.asme.org/dynamicsystems/article-pdf/136/5/051015/6116911/ds\_136\_05\_051015.pdf},
}

@article{Bhattacharjee2018,
  author = {Sourav Bhattacharjee and D. Schnieders},
  title = {AUTONOMOUS DRIFTING {RC} CAR WITH REINFORCEMENT LEARNING},
  journal = {interim report},
  year = {2018},  
  url  = {https://i.cs.hku.hk/fyp/2017/fyp17014/docs/InterimReport.pdf}

}

@article{Spielberg2019,
author = {Nathan A. Spielberg  and Matthew Brown  and Nitin R. Kapania  and John C. Kegelman  and J. Christian Gerdes },
title = {Neural network vehicle models for high-performance automated driving},
journal = {Sci. Robot.},
volume = {4},
number = {28},
pages = {eaaw1975},
year = {2019},
doi = {10.1126/scirobotics.aaw1975},
URL = {https://www.science.org/doi/abs/10.1126/scirobotics.aaw1975},
eprint = {https://www.science.org/doi/pdf/10.1126/scirobotics.aaw1975},
}

@article{Yin2020,
    author = {Yin, Yuming and Li, Shengbo Eben and Li, Keqiang and Yang, Jue and Ma, Fei},
    title = {Self-learning drift control of automated vehicles beyond handling limit after rear-end collision},
    journal = {Transp. Saf. and Environ.},
    volume = {2},
    number = {2},
    pages = {97-105},
    year = {2020},
    month = {05},
    issn = {2631-4428},
    doi = {10.1093/tse/tdaa009},
    url = {https://doi.org/10.1093/tse/tdaa009},
}

@article{Wurman2021,
  author    = {Wurman, Peter R. and others},
  title     = {Outracing champion Gran Turismo drivers with deep reinforcement learning},
  journal   = {Nature},
  volume    = {602},
  pages     = {223--228},
  year      = {2022},
  doi       = {10.1038/s41586-021-04357-7},
  url       = {https://doi.org/10.1038/s41586-021-04357-7}
}

@article{Orgovan2021, title={Autonomous Drifting Using Reinforcement Learning}, volume={49}, url={https://pp.bme.hu/tr/article/view/18581}, DOI={10.3311/PPtr.18581}, abstractNote={&lt;p&gt;Autonomous vehicles or self-driving cars are prevalent nowadays, many vehicle manufacturers, and other tech companies are trying to develop autonomous vehicles. One major goal of the self-driving algorithms is to perform manoeuvres safely, even when some anomaly arises. To solve these kinds of complex issues, Artificial Intelligence and Machine Learning methods are used. One of these motion planning problems is when the tires lose their grip on the road, an autonomous vehicle should handle this situation. Thus the paper provides an Autonomous Drifting algorithm using Reinforcement Learning. The algorithm is based on a model-free learning algorithm, Twin Delayed Deep Deterministic Policy Gradients (TD3). The model is trained on six different tracks in a simulator, which is developed specifically for autonomous driving systems; namely CARLA.&lt;/p&gt;}, number={3}, journal={Periodica Polytechnica Transp. Eng.}, author={Orgován, László and Bécsi, Tamás and Aradi, Szilárd}, year={2021}, pages={292–300} }

@INPROCEEDINGS{Jiang2021,
  author={Jiang, Yan and Xu, Xin and Zhang, Xinglong and Huang, Junwen and Gao, Shuai},
  booktitle={2021 China Automat. Congr. (CAC)}, 
  title={Reinforcement Learning-based Drift Control for Autonomous Vehicles}, 
  year={2021},
  volume={},
  number={},
  pages={4508-4513},
  doi={10.1109/CAC53003.2021.9727514}}

@INPROCEEDINGS{Domberg2022,
  author={Domberg, Fabian and Wembers, Carlos Castelar and Patel, Hiren and Schildbach, Georg},
  booktitle={2022 IEEE Int. Conf. on Robot. and Automat. (ICRA)}, 
  title={Deep Drifting: Autonomous Drifting of Arbitrary Trajectories using Deep Reinforcement Learning}, 
  year={2022},
  volume={},
  number={},
  pages={7753-7759},
  doi={10.1109/ICRA46639.2022.9812249}}

@article{Toth2023,
title = {Tabular {Q}-learning Based Reinforcement Learning Agent for Autonomous Vehicle Drift Initiation and Stabilization},
journal = {IFAC-PapersOnLine},
volume = {56},
number = {2},
pages = {4896-4903},
year = {2023},
note = {22nd IFAC World Congress},
issn = {2405-8963},
doi = {https://doi.org/10.1016/j.ifacol.2023.10.1261},
url = {https://www.sciencedirect.com/science/article/pii/S2405896323016658},
author = {Szilárd H. Tóth and Ádám Bárdos and Zsolt J. Viharos}
}

@article{Leng2023,
  author    = {Bo Leng and Yuhang Yu and Ming Liu and others},
  title     = {Deep reinforcement learning-based drift parking control of automated vehicles},
  journal   = {Sci. China Technological Sciences},
  volume    = {66},
  number    = {5},
  pages     = {1152--1165},
  year      = {2023},
  doi       = {10.1007/s11431-022-2273-5},
  url       = {https://doi.org/10.1007/s11431-022-2273-5}
}

@ARTICLE{Weber2023,
  author={Weber, Trey P. and Gerdes, J. Christian},
  journal={IEEE Trans. Intell. Veh.}, 
  title={Modeling and Control for Dynamic Drifting Trajectories}, 
  year={2024},
  volume={9},
  number={2},
  pages={3731-3741},
  doi={10.1109/TIV.2023.3340918}}

@INPROCEEDINGS{Domberg2024,
  author={Domberg, Fabian and Barkow, Ben and Schildbach, Georg},
  booktitle={AmEC 2024 – Automot. meets Electro. and Control; 14. GMM Symp.}, 
  title={Vision-based Autonomous Trajectory Drifting using Deep Reinforcement Learning}, 
  year={2024},
  volume={},
  number={},
  pages={47-52},
  doi={}}

@INPROCEEDINGS{Hoshino2024,
  author={Hoshino, Hikaru and Li, Jiaxing and Menon, Arnav and Dolan, John M. and Nakahira, Yorie},
  booktitle={2024 IEEE 27th Int. Conf. on Intell. Transp. Syst. (ITSC 2024)}, 
  title={Autonomous Drifting Based on Maximal Safety Probability Learning}, 
  year={2024},
  volume={},
  number={},
  pages={3930-3935},
  doi={10.1109/ITSC58415.2024.10919509}}

@ARTICLE{Zhao_2025_drift,
  author={Zhao, Shiyue and Zhang, Junzhi and Masoud, Neda and Jiang, Yuhong and Huang, Heye and Liu, Tao},
  journal={IEEE Transactions on Industrial Electronics}, 
  title={Drift Cornering Control and Real-Vehicle Deployment for Electric Vehicles}, 
  year={2025},
  volume={72},
  number={12},
  pages={13509-13520},
  doi={10.1109/TIE.2025.3563719}}

@ARTICLE{Zhao2024_Adaptive,
  author={Zhao, Shiyue and Zhang, Junzhi and He, Chengkun and Hou, Xiaohui and Huang, Heye},
  journal={IEEE Trans. Ind. Electron.}, 
  title={Adaptive Drift Control of Autonomous Electric Vehicles After Brake System Failures}, 
  year={2024},
  volume={71},
  number={6},
  pages={6041-6052},
  doi={10.1109/TIE.2023.3294594}}

@article{WANG2024,
title = {Learning autonomous race driving with action mapping reinforcement learning},
journal = {ISA Trans.},
volume = {150},
pages = {1-14},
year = {2024},
issn = {0019-0578},
doi = {https://doi.org/10.1016/j.isatra.2024.05.010},
url = {https://www.sciencedirect.com/science/article/pii/S0019057824002143},
author = {Yuanda Wang and Xin Yuan and Changyin Sun}
}

@INPROCEEDINGS{Bertipaglia_2024_model,
  author={Bertipaglia, Alberto and Tavernini, Davide and Montanaro, Umberto and Alirezaei, Mohsen and Happee, Riender and Sorniotti, Aldo and Shyrokau, Barys},
  booktitle={2024 IEEE International Conference on Advanced Intelligent Mechatronics (AIM)}, 
  title={Model Predictive Contouring Control for Vehicle Obstacle Avoidance at the Limit of Handling Using Torque Vectoring*}, 
  year={2024},
  volume={},
  number={},
  pages={1468-1475},
  doi={10.1109/AIM55361.2024.10637113}}

@inproceedings{Djeumou2024,
title={One Model to Drift Them All: Physics-Informed Conditional Diffusion Model for Driving at the Limits},
author={Franck Djeumou and Thomas Jonathan Lew and Nan Ding and Michael Thompson and Makoto Suminaka and Marcus Greiff and John Subosits},
booktitle={8th Annual Conf. on Robot Learn.},
year={2024},
}

@InProceedings{Broadbent2024,
author="Broadbent, Nicholas Drake
and Weber, Trey
and Mori, Daiki
and Gerdes, J. Christian",
title="Neural Network Tire Force Modeling for Automated Drifting",
booktitle="16th Int. Symp. on Adv. Vehicle Control",
year="2024",
publisher="Springer Nature Switzerland",
address="Cham",
pages="378--384",
isbn="978-3-031-70392-8"
}

@article{Djeumou2025,
  title={Reference-Free Formula Drift with Reinforcement Learning: From Driving Data to Tire Energy-Inspired, Real-World Policies},
  author={Franck Djeumou and Michael Thompson and Makoto Suminaka and John Subosits},
  journal={ArXiv},
  year={2024},
  volume={abs/2410.20990},
  url={https://api.semanticscholar.org/CorpusID:273654870}
}

\balance

\end{document}